\documentclass[10pt,journal,letterpaper,compsoc]{IEEEtran}

\usepackage{iccv}
\usepackage{graphicx}
\usepackage{amsmath}
\usepackage{amssymb}
\usepackage{floatflt}
\usepackage{subfigure}
\usepackage{color}
\usepackage{algorithm}
\usepackage{algorithmic}

\graphicspath{{figs/}{pics/}{pics//face/}}
\DeclareGraphicsExtensions{.jpg,.pdf,.mps,.png}

\iccvfinalcopy 

\def\iccvPaperID{404} 

\ificcvfinal\fi

\def\argmin{\mathop{\rm argmin}}
\def\RR{\mathbb R}

\title{Hierarchical Object Parsing from Structured Noisy Point Clouds}

\author{Adrian Barbu\thanks{A. Barbu is with the Department of Statistics, Florida State University, Tallahassee, Florida 32306, USA,
Phone: 850-980-2516, Fax: 850-644-5271, Email: abarbu@stat.fsu.edu.} }

\begin{document}
\maketitle
\begin{abstract}
Object parsing and segmentation from point clouds are challenging tasks because the relevant data is available only as thin structures along object boundaries or other features, and is corrupted by large amounts of  noise. To handle this kind of data, flexible shape models are desired that can accurately follow the object boundaries. Popular  models such as Active Shape and Active Appearance models lack the necessary flexibility for this task, while recent approaches such as the Recursive Compositional Models make model simplifications in order to obtain computational guarantees. This paper investigates a hierarchical Bayesian model of shape and appearance in a generative setting. The input data is explained by an object parsing layer, which is a deformation of a hidden PCA  shape model with Gaussian prior. The paper also introduces a novel efficient inference algorithm that uses informed data-driven proposals to initialize local searches for the hidden variables. Applied to the problem of object parsing from structured point clouds such as edge detection images, the proposed approach obtains state of the art parsing errors on two standard datasets without using any intensity information.
\end{abstract}

\section{Introduction}

Object parsing and segmentation are important problems with many applications in computer vision and medical imaging. While object segmentation is only directed towards labeling the object pixels, object parsing is aimed at identifying the object parts such as head, body, legs, etc. 

The object parsing problem presents challenges in both modeling and computing. It is difficult to find accurate probability or energy models that have high probability on correct object parsings of the input image and low probability everywhere else. Moreover, it is difficult to design inference algorithms that find the correct object parsing from the image in a reasonable amount of time. Most inference problems for non-tree Markov Random Field (MRF) based models are NP-hard (proved in \cite{boykov2001fast} for the Potts model) so an exact solution cannot be expected in polynomial time. 

Because of the computational challenges, different trade-offs between model accuracy and computational feasibility have been made in previous works. 

The Active Shape \cite{cootes1995active} and Active Appearance Models \cite{cootes2001active} use a simplified object representation using Principal Component Analysis (PCA) and use local information to search for a solution. The Active Shape Models (ASM) contain only a PCA shape model and alternate one step that searches for local boundary evidence on the shape normals with another step that reprojects the evidence onto the PCA hyperplane, until convergence. The trade-off made by the ASM is the local search for the solution, based on partial image information existent on the shape normals. Because of this trade-off, the result depends on initialization. 

\vskip -2mm
\begin{figure}[ht]
\centering
\includegraphics[width=6.cm]{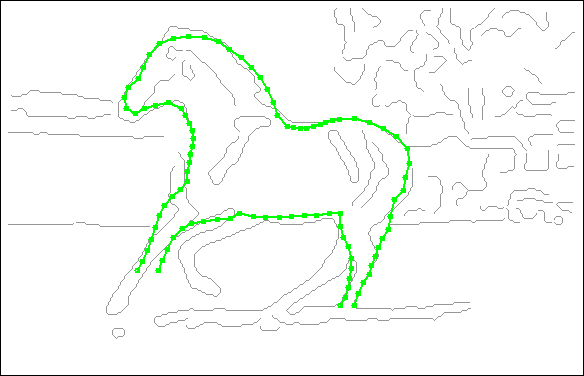}
\vskip -0mm
\caption{Motivation for the hierarchical model. A shape described by PCA (shown with green dots) is not flexible enough to accurately follow the object boundary, but can serve as a backbone to limit the variability of the model.}
\label{fig:example}
\end{figure}
\vskip -3mm
\begin{figure*}[ht]
\centering
\includegraphics[width=5.7cm]{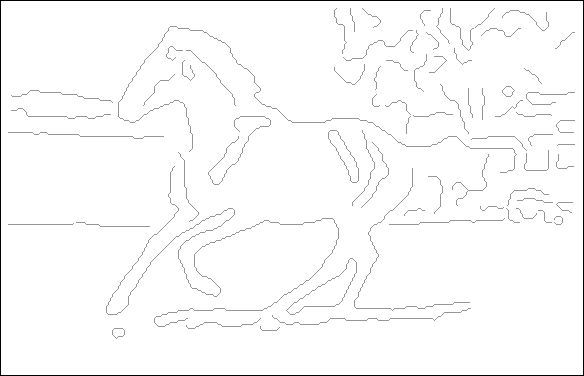}
\includegraphics[width=5.7cm]{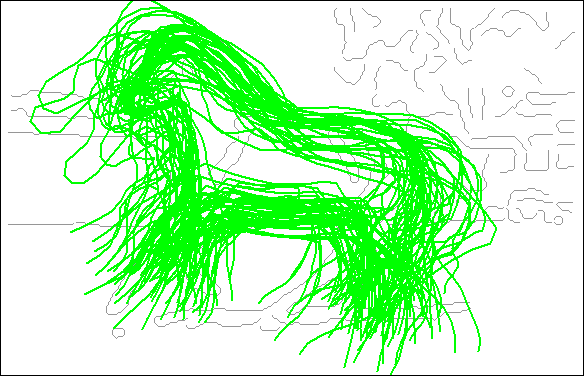}
\includegraphics[width=5.7cm]{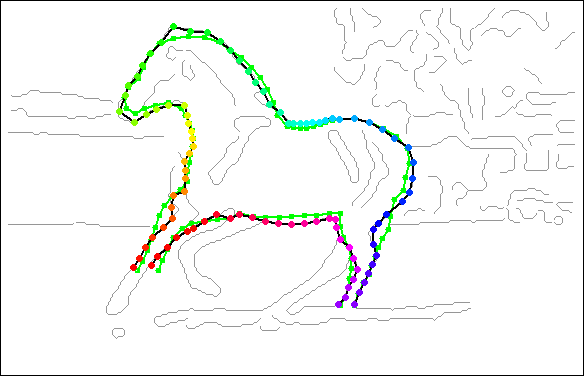}
\caption{Our approach starts by tracing the points into chains (left), finds data-driven PCA candidates (middle) that are used to initialize local optimizations of the model parameters. The parameters of lowest energy give the parsing result (right, black with colored dots) and associated PCA shape (right, green).}
\label{fig:segexample}
\vspace{-6mm}
\end{figure*}The Active Appearance Models (AAM) also model the object appearance by PCA and employ a trained iterative algorithm that deforms an initial shape until convergence, guided by the image. The AAM uses more image information than the ASM, but the algorithm is still greedy, obtaining a local optimum that is dependent on initialization. In \cite{IMM2000-0124,Stegmann2003tmi} the radius of attraction of the AAM modes was used together with a measure of error based on sum of squares to obtain a globally optimal solution of the AAM.

Another limitation of the ASM/AAM models is the rigidity, as a low dimensional PCA shape cannot accurately describe the shape variability existent in real images, and is limited only to the main deformations. This is illustrated in Figure \ref{fig:example}, where a 10-dimensional PCA shape shown in green cannot accurately follow the horse boundary and is off by a few pixels around the ears, back, legs, etc. 

Many other previous works \cite{li2009robust,zhu2009learning} make trade-offs in the model in order to be able to cast it into a class for which efficient inference algorithms exist, and will be discussed in more detail in the next section. 

The approach introduced in this paper tries to avoid making any modeling compromises, at the expense of having no off-the-shelf algorithm for inference. Instead, we introduce a novel inference algorithm that searches for many local optima using data-driven techniques and is more likely to find a solution close to the global optimum than a local optimization algorithm. The quantitative experiments will tell us how successful this strategy was in obtaining good parsing results. We observe that the proposed  model together with the suboptimal algorithm can obtain state of the art object parsing results without using any intensity information. This finding is in line with our previous work \cite{barbu2009training} where a properly trained model compensated for a fast suboptimal algorithm for image denoising.

The contributions of the paper are the following:

First, it presents a hierarchical generative model that represents the object shape as a MRF-based deformation from a PCA backbone, obtaining a more accurate boundary delineation than with the PCA model alone. The shape model can be sampled if desired and used for numerical integration or to compute marginal statistics. The generative model also contains a data term that connects the image information with the shape model. Due to the high accuracy of the shape description, this model can be used for object parsing from point clouds (such as those obtained from edge detection), where the data information is one pixel wide. 

Second, it presents an optimization algorithm for finding a strong optimum for the hierarchical model. The algorithm, illustrated in Figure \ref{fig:segexample}, uses a data-driven set of PCA candidates to initialize local searches for optimizing a unique energy based on the MRF deformation and PCA parameters. 

Third, it presents an evaluation of the proposed approach to parsing horses \cite{borenstein2002class}, cows  \cite{leibe2004combined} and faces \cite{IMM20020922} from point clouds obtained from edge detection. The evaluation revealed that the proposed algorithm obtains state of the art results on two of the datasets without using any intensity information.

The proposed shape model and inference algorithm could in principle be adapted for other object parsing and segmentation applications by using appropriate data terms, or even by employing more accurate backbone shape models than the PCA.

\section{Related Work}

Matching point clouds dates back to the Softassign \cite{rangarajan1997softassign} and the Robust Point Matching Algorithm \cite{chui2003new}. These methods are capable of finding correspondences and transformations between unordered point clouds. However, for the problem of object parsing, their shape model can be considered a template plus deformation, which might not be an accurate enough model for real objects. It would be interesting to see how these methods perform in finding real objects from real edge detection images such as those presented in this paper.

An ASM based approach containing a dynamic programming step similar to the one in this paper was proposed in \cite{behiels1999active}. However, the appearance model from \cite{behiels1999active} is based on learned intensity profiles, thus it relies heavily on the intensity information. Moreover, the approach depends on an initialization step and does not have a hierarchical energy formulation as the one proposed in this paper.

The Oriented ASM \cite{liu2009oriented} also uses dynamic programming to find the minimum cost boundary. The cost is based on intensity profiles perpendicular to the boundary at the landmark points and a cost along a contour obtained by the live wire method \cite{falcao1998user} between consecutive landmark points. The regularized PCA shape is used to constrain the obtained boundary, while in our approach it is part of the energy formulation, allowing deformations from the PCA shape.

A probabilistic model of shape and appearance was used in the Constrained Local Models (CLM) \cite{cristinacce2008automatic}. The appearance was modeled using local AAM  templates at a number of keypoints on the object to be parsed. A Bayesian model combined a measure of appearance similarity of the local image templates with a global PCA shape prior similar to the one used in this paper. A local optimum of the joint model was obtained using the simplex method.

A refinement of the CLM is the cascade of Combined Shape Models (c-CSM) \cite{tresadern2009combining}. This approach alternates two steps, one maximizing a posterior probability $p(Y|I)$ of an approximate solution $Y$ and one of obtaining a regularized solution $X$ from $Y$, which maximizes $p(X|Y)$. Using this terminology, our approach optimizes a single energy function $E(X,Y)$ that has a likelihood term $p(I|Y)$ and a prior $p(Y|X)$ based on the regularized shape $X$, and a prior $p(X)$. In c-CSM the final result is the regularized shape $X$ while in our approach the final result is $Y$. Our inference algorithm is also different from the c-CSM in that we use data-driven candidates.

A robust hierarchical shape model was constructed for multi-view car alignment \cite{li2009robust}. The model is a probabilistic PCA, which is a PCA model plus i.i.d Gaussian deformation of each vertex. It allows large deformations of the observed shape points and can also handle missing points due to occlusion or failures of the part detectors. Similar to our work, shape candidates are constructed from partial information obtained from part detectors to initialize a local search. However, these candidates have been directly generated using RANSAC, as the correspondence between the part detections and the model points was known. In the object parsing from point clouds, the correspondence between input points and the object points is not known and the fraction of outliers is usually higher than $90\%$, making RANSAC computationally prohibitive (a computation using eq. (13) from \cite{li2009robust} gives about $4\cdot 10^5$ candidates required for $99\%$ certainty).

The Recursive Compositional Models (RCM) \cite{zhu2009learning} represents the object shape in a hierarchical fashion using multiple levels of rotation-invariant models based on triplets of elements. The first level elements are the detected image edges, while the elements for each subsequent level are summaries of the triplets from the previous level. This hierarchical model allows inference by a version of dynamic programming with pruning. Through the hierarchy, the model enforces long range interactions between the shape elements, but at the same time some desired short-range interactions are missing.
In contrast, our hierarchical model represents the shape using a PCA model plus MRF deformations along the normals, with the PCA model providing the long range interactions while the MRF allowing for smooth deformations. Because of the high connectivity of our proposed model, exact inference algorithms based on dynamic programming are not applicable. Instead, we propose a smart search algorithm that makes local searches at  a number of locations dictated by a bottom-up data-driven process. The advantage of our approach is the simplicity of the model, that can be easily learned from training examples. Our evaluation shows that the errors obtained by our approach are similar to the RCM on two of the three datasets, without using any image intensity information.

The knowledge based segmentation \cite{besbes2009shape} uses a shape prior based on pairwise cliques between the shape points and a primal-dual algorithm for inference. In contrast, our framework uses a PCA-based model that cannot be decomposed in pairwise cliques and could in principle be extended to work with non-linear shape models.


Torresani et al, \cite{torresani2003learning} model the shape as a rigid transformation plus PCA, without the MRF deformation from our formulation.

Interactive Object Segmentation with Graph Cuts \cite{freedman2005interactive} imposes a shape prior on a Graph Cut energy. However, the shape prior is based on a template with similarity transformation without any deformation and the Graph Cut energy is on pixels, so no object parsing or aligned boundary is obtained. This work has been extended with a Kernel PCA shape prior \cite{malcolm2007graph}, but still depends on manual initialization and obtains just a segmentation without boundary alignment. In contrast, our method obtains object parsing and boundary alignment, hence not only the object boundary but also the object parts are obtained.

The work of Ren et al, \cite{ren2006cue} is targeted to object boundary detection. If does not obtain a clear object segmentation or a parsing into object parts. Furthermore, it uses both edge and gradient information as input data.

Felzenszwalb and Schwartz, \cite{felzenszwalb2007hierarchical} use a shape tree as a model and focus on shape matching and retrieval, without evaluating the parsing error.

Zhu et al, \cite{zhu2007untangling} use a circularity measure to find cycles with good continuation in edge detection images. However, it does not have any global shape model so it addresses a different problem than ours.

The Active Skeleton \cite{bai2009active} uses a skeleton-based shape model to detect objects from edge detection images. Even though in principle the method could be used for object parsing, it has not been evaluated for this purpose.

Groups of nearby contour segments are used in \cite{ferrari2007groups,shottonieee} to construct features for object detection. Our paper uses similar contour fragments to construct bottom-up data driven candidates for searching the shape space. The features from \cite{ferrari2007groups,shottonieee} could be further used to obtain better discriminative object models. Currently we use a generative model, with some parameters trained in a discriminative manner.

The part-based constellation model from \cite{stark2010shape} uses an extension of the contour segment network from \cite{ferrari2007groups} to construct object parts and a Metropolis-Hastings stochastic algorithm for inference. However, the method is used for object detection, where the precise location of the boundary is not as important as in segmentation or parsing.

Another closely related work is the unsupervised learning of shape models \cite{ferrari2010images}, which uses pairs of adjacent contours \cite{ferrari2007groups} as features and a voting scheme to find the object parameters. A separate deformation step is then performed using Thin Plate Splines. In contrast, the inference algorithm from our work optimizes a single criterion that combines the shape and deformation into a single hierarchical model. Moreover, our work is aimed towards object parsing, whereas \cite{ferrari2010images} is used for object and boundary detection.

The Active Basis Model \cite{wu2009learning} can obtain a sketch of an object using Gabor filters and has been successfully used for object detection. However, it has not been evaluated for object parsing or segmentation, and the sketch elements are not subject to a smoothness prior.

Our approach is inspired by \cite{fernandes2008real}, where an efficient version of the Hough transform for line detection is obtained by voting at locations given by least squares line fitting of clusters of approximately collinear pixels. 

Generating candidates based on partial information is similar to the beta channel from \cite{yang2010evaluating}, where partially occluded faces are detected by combining eye, nose and mouth detections.

\section{A Hierarchical Approach to Object Parsing}

We propose a hierarchical generative model with two levels of hidden variables that need to be inferred from the input data. 
The first level $C$ is the actual object parsing while the second level is a PCA shape model that limits the degree of variability of the first level. 

The PCA shape is controlled by variables $(A,\beta)$ consisting of a similarity transformation $A$ and the PCA coefficients $\beta\in \RR^p$. 
We abuse the notation by denoting $A$ as both the transformation parameters $A=(u,v,s,\theta)$, with rotation $\theta$, translation $(dx,dy)$ and scale $s$, and the actual transformation
\[
A(x,y)=(sx \cos \theta +sy\sin \theta+u, -sx\sin \theta +sy\cos \theta+v) 
\]
The PCA shape is 
\vspace{-1mm}
\begin{equation}
S(A,\beta)=A(\mu_x+P_x\beta,\mu_y+P_y\beta)=(S_1,...,S_N)'\label{eq:pca}
\vspace{-1mm}
\end{equation}
 where $\mu=(\mu_x,\mu_y)\in \RR^N\times \RR^N$ is the mean shape and $P=(P_x,P_y),P_x,P_y\in {\cal M}_{N,p}$ are the PCA eigenvectors (${\cal M}_{N,p}$ being the space of $N\times p$ real matrices). 
\vskip -4mm
\begin{figure}[ht]
\centering
\includegraphics[width=7.5cm]{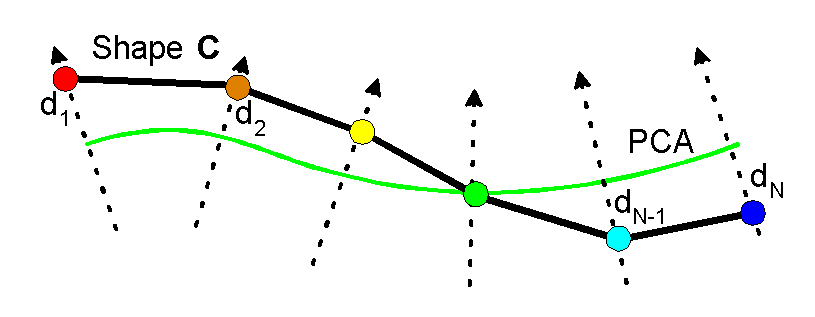}
\vskip -1mm
\caption{The shape $C$ is obtained as a deformation ${\bf d}=(d_1,...,d_N)$ of a PCA shape $(A,\beta)$ along the normals.}
\label{fig:asm}
\end{figure}
\vskip -4mm

The representation is illustrated in Figure \ref{fig:asm}. The shape $C$ (black) is obtained from the PCA shape (green) using a vector of displacements ${\bf d}=(d_1,...,d_N)\in [-d_{max},d_{max}]^N$ along the normals to shape. More exactly, the shape $C=C({\bf d})$ consists of a number of line segments $\overline{C_i,C_{i+1}}$ where $C_i=S_i+n_i d_i,i=1,...,N$ and $n_i$ is the normal to the PCA shape at $S_i$.

\subsection{The Hierarchical Generative Model}

\begin{figure}[h]
\centering
\includegraphics[width=8.5cm]{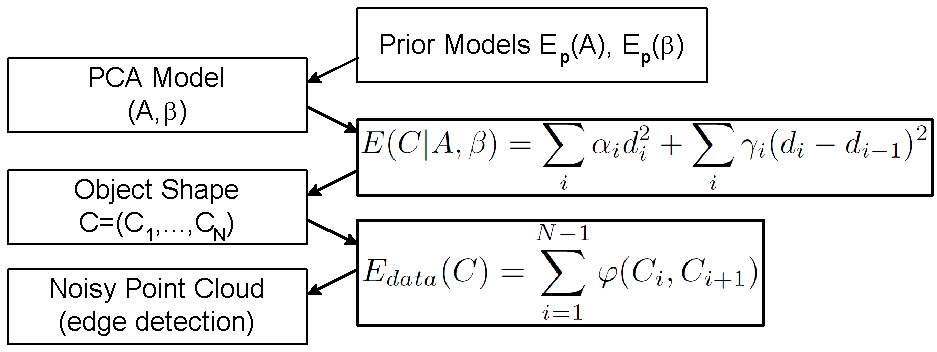}
\vskip -1mm
\caption{Diagram of the proposed hierarchical model.}
\label{fig:model}
\vspace{-6mm}
\end{figure}

The model can be represented either as a probability or an energy. 
For simplicity, we use an energy formulation of the model, illustrated in Figure \ref{fig:model},
\vspace{-1mm}
\begin{equation}
E(C,A,\beta)\hspace{-0.5mm}=\hspace{-0.5mm}E_{data}(C)\hspace{-1mm}+E_{shape}(C,\beta|A)+\hspace{-1mm}E_{p}(A),\label{eq:model}
\vspace{-1mm}
\end{equation}
containing a data term $E_{data}(C)$ that relates the input data with the parsing result $C$, a shape deformation term $E_{shape}(C,\beta|A)$ and a  prior $E_p(A)$ on the possible transformations $A$. 
The data term  $E_{data}(C)$ is application specific and is based on the exact location of the shape $C=(C_1,...,C_N)$.  When the input data consists of noisy point clouds such as edge detection, the input points are traced into point chains based on the 8-neighborhood.
The data term $E_{data}(C)$ encourages consecutive points $C_i,C_{i+1}$ to be on the same point chain: 
\vspace{-2mm}
\begin{equation}
E_{data}(C)=\sum_{i=1}^{N-1}\varphi(C_i,C_{i+1})
\vspace{-2mm}
\end{equation}
where $\varphi(C_i,C_{i+1})=-\delta$ if and only if $C_i,C_{i+1}$ are on the same point chain and $\varphi(C_i,C_{i+1})=0$ otherwise.

The shape term
\vspace{-1mm}
\begin{equation}
E_{shape}(C,\beta|A)=E(C|A,\beta)+E_{p}(\beta)\label{eq:shape}
\vspace{-1mm}
\end{equation}
consists of  a deformation term $E(C|A,\beta)$ that connects the parsed shape $C$ with the underlying PCA model with parameters $(A,\beta)$, and a prior $E_{p}(\beta)$ on the PCA coefficients $\beta$ (which are assumed to be independent of the transformation $A$).

The deformation term $E(C|A,\beta)$ is a Gaussian MRF \cite{amit1991structural} that encourages the curve (or curves) to be parallel and close to the PCA shape
\vspace{-1mm}
\begin{equation}
E(C|A,\beta)=\sum_i \alpha_i d_i^2+ \sum_i \gamma_i (d_i-d_{i-1})^2
\vspace{-2mm}
\end{equation}
and is defined in terms of the displacements $d_i$ of the curve points $C_i$ from the corresponding PCA shape points $S_i$. 
The coefficients $\alpha_i,\gamma_i$ represent the amount of penalty for the deformation at different points along the shape. 

If the object contains multiple contour segments, they are concatenated into a single contour $C$ and the coefficients $\gamma_i$ connecting points on different contour segments will be zero. Similarly, there will be no data term $\phi(C_i,C_{i+1})$ between the contour segments.

In our applications all $\alpha_i$ have the same value $\alpha_i=\alpha$ and similarly $\gamma_i=\gamma$ (except those connecting different contour segments which are 0). This simplification could result in a decreased performance. For example the $\alpha_i$ for points on the horse head could have smaller values because there is more variability for those points.

The prior $E_p(\beta)$ on the PCA parameters is a Gaussian prior based on the PCA eigenvalues $\lambda_i$
\vspace{-2mm}
\begin{equation}
E_p(\beta)=\rho\sum_{i=1}^N \frac{\beta_i^2}{\lambda_i}
\vspace{-2mm}
\end{equation}

The prior $E_p(A)$ for $A=(u,v,s,\theta)$ forces the scale and rotation within a range and discourages translations away from the image center $(x_c,y_c)$:
\vspace{-2mm}
\begin{equation}
E_p(A)=\hspace{-1mm}\begin{cases} \infty \text{   if $s\not \in [s_{min},s_{max}]$  or $|\theta|\hspace{-1mm}>\hspace{-1mm} \theta_{max}$}\\
r |u-x_c|+r |v-y_c| \quad \text {else}
\end{cases}\label{eq:posprior}
\vspace{-2mm}
\end{equation}

 The model parameters $\Theta=(\alpha,\gamma,\delta,\rho,r)$ are learned in a supervised manner on the training set through a procedure described in Section \ref{sec:learn}.

One advantage of the generative model described in eq. \eqref{eq:shape} is that one could easily obtain samples from the shape model $E(C|A,\beta)$, by sampling the PCA coefficients $\beta$ from the Gaussian prior $\beta \sim \frac{1}{Z_1}\exp (-E_p(\beta))$ and the deformation field ${\bf d}$ from the Gaussian MRF ${\bf d}\sim   \frac{1}{Z_2}\exp (-\sum_i \alpha_i d_i^2+ \sum_i \gamma_i (d_i-d_{i-1})^2)$. 
\vskip -4mm
\begin{figure}[htb]
\centering
\includegraphics[width=3.8cm]{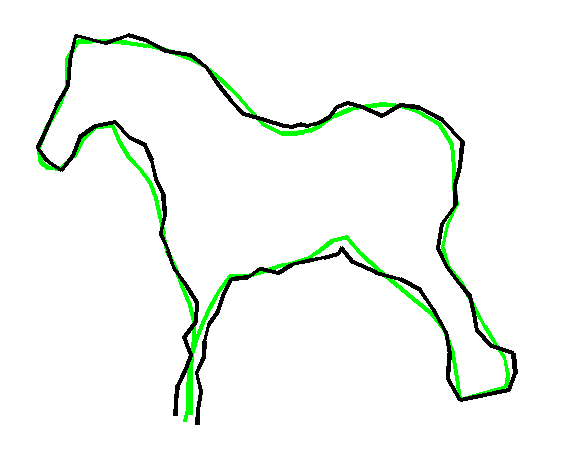}
\includegraphics[width=3.8cm]{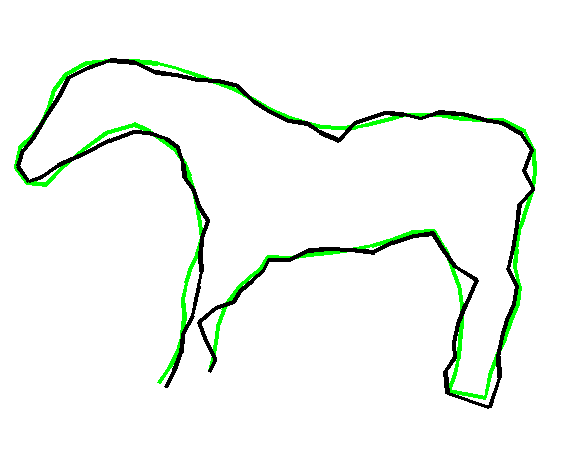}
\includegraphics[width=3.8cm]{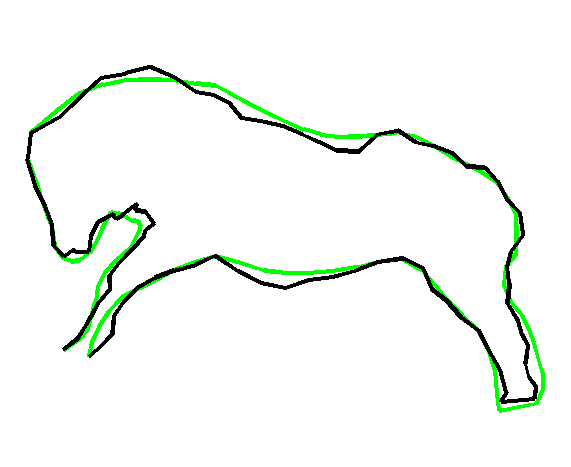}
\includegraphics[width=3.8cm]{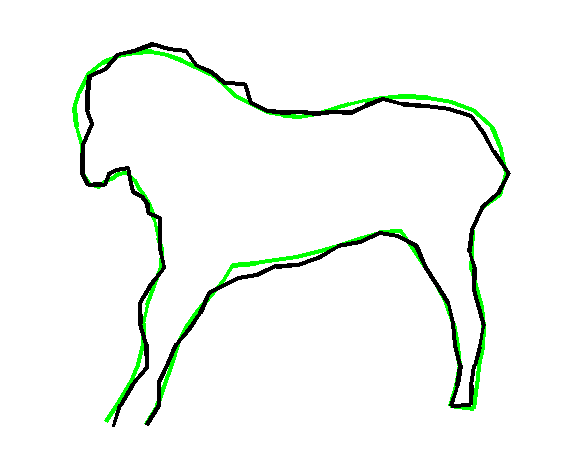}
\vskip -3mm
\caption{Sample shapes from the shape model \eqref{eq:shape} with the parameters of the learned horse model. PCA shape (green) and sampled shape (black).}
\label{fig:sampleshapes}
\end{figure}
\vskip -4mm
In Figure \ref{fig:sampleshapes} are shown a few samples from the learned horse model  $E(C|A,\beta)$ with the PCA shape $S$ shown in green and the sampled shape $C$ in black.

\subsection{Inference Algorithm}

Finding the object parsing $C$ and the PCA parameters $(A,\beta)$ is a nontrivial optimization problem. 
\vskip -3mm
\begin{figure}[htb]
\centering
\includegraphics[width=3.5cm]{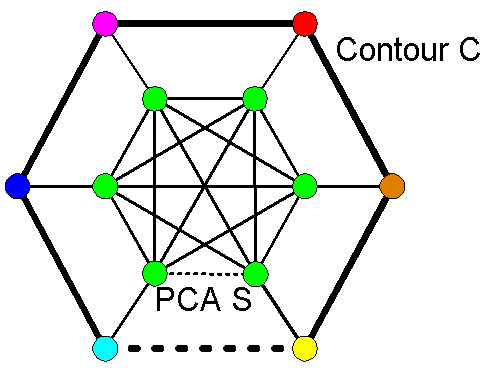}
\vskip -0.1mm
\caption{\label{fig:MRF} The MRF interaction between the hidden variables has a large clique (between the PCA points S showed in green) and a number of binary cliques.}
\vspace{-3mm}
\end{figure}
The hidden variables are connected through a MRF, as illustrated in Figure \ref{fig:MRF}. The PCA points (green) form a large fully connected clique in the MRF energy, and each contour point is  connected to a PCA point and with its neighbors through pairwise cliques. The node labels represent the positions of the corresponding points in the image.

There exist recent advances in optimization for MRF energies with higher order cliques, such as \cite{kohli2009robust} extending Graph Cuts \cite{boykov2001fast} or \cite{komodakis2009beyond,jojic2010accelerated} based on Dual Decomposition. However, one could not even exhaust all possible combinations of labels on the nodes of the large clique, because it is computationally unfeasible even when the nodes have binary labels. For example, the large clique has 96 nodes for the horse parsing task.

We adopt a different strategy instead. If the PCA parameters $(A,\beta)$ are known, the parsing $C$ is uniquely determined by the displacement vector ${\bf d}=(d_1,...,d_N)$, hence $C=C({\bf d})$. In this case finding the optimal $C({\bf d})$ is equivalent to finding $\bf d$ that minimizes $E(C({\bf d}),A,\beta)=E({\bf d})$. This can be done efficiently by dynamic programming, due to the additive nature of the model when the PCA shape $(A,\beta)$ is fixed.

\vspace{-2mm}
{\small
\[E({\bf d})=\sum_{i=1}^{N-1}\varphi(C_i,C_{i+1})+\sum_i \alpha_i d_i^2+ \sum_i \gamma_i (d_i-d_{i-1})^2+ct.
\vspace{-2mm}
\]
}
If the parsing $C$ is fixed, an approximate minimum of $E(C,A,\beta)$ can be obtained by least square fitting of the PCA shape parameters $(A,\beta)$.

Therefore, if the PCA shape parameters $(A,\beta)$ are initialized close to their optimal values, an approach that alternates the above two steps, namely the computation of the parsing $C$ by dynamic programming and the estimation of the PCA parameters $(A,\beta)$ by least squares, will converge to an approximate local optimum of $E(C,A,\beta)$ in a few iterations. This approach is similar in spirit to the Active Shape Model, with the difference that a smooth contour $C$ is found by optimization in our method instead of finding data evidence on each normal independently as the ASM does.
\begin{algorithm}[htb]
   \caption{{\bf Optimization Algorithm}}
   \label{alg:inference}
\begin{algorithmic}
   \STATE {\bfseries Input:} Noisy point cloud e.g. edge detection image, PCA candidates $(A_i,\beta_i),i=1,...,N^{cand}$.
   \STATE {\bfseries Output:} Near-optimal hidden variables $(\hat C,\hat A,\hat \beta)$.
	\FOR { $i=1$ {\bfseries to} $N^{cand}$}   
	 \FOR { $j=1$ {\bfseries to} $N_{iter}$}
	    \STATE Find displacement vector $\bf d$ using dynamic programming 
	\vspace{-2mm}
			\begin{equation}
			{\bf d}=\argmin_{\bf d}	E_{data}(C({\bf d}))+E({\bf d}|A_i,\beta_i).
\vspace{-2mm}
			\end{equation}
   		\STATE Refit $(A_i,\beta_i)$ by least squares on $C({\bf d})$
   \ENDFOR
   \STATE Obtain $C_i=C({\bf d})$.
   \ENDFOR
   \STATE Find $j=\argmin_i E(C_i,A_i,\beta_i)$ 
   \STATE Obtain	$(\hat C,\hat A,\hat \beta)=(C_j,A_j,\beta_j)$
\end{algorithmic}
\end{algorithm}

We will use a data-driven approach described in Section \ref{sec:cg} to obtain a number of PCA candidate shapes $(A_i,\beta_i), i=1,...,N^{cand}$ for initialization of the local search described above.  The final solution is obtained as the lowest energy configuration $(C,A,\beta)$ among the $N^{cand}$ local optima obtained. The whole optimization algorithm is described in Algorithm \ref{alg:inference}.

As the model energy \eqref{eq:model} is just an approximation of the true object shape model, it is possible that other ways to combine the candidates such as weighted averaging \cite{sofka2010multiple} might be better than choosing the lowest energy one. This is subject to further investigation.

\subsection{PCA Candidate Generation} \label{sec:cg}

The PCA shape candidates $(A_i,\beta_i), i=1,...,N^{cand}$ are obtained by matching one or more contour fragments to parts of the PCA model. The contour fragments are similar to \cite{ferrari2007groups,shottonieee} and are obtained in a preprocessing step described in Section \ref{sec:preprocessing} below.

An initial set of PCA candidates can obtained from one contour fragment, as described in Section \ref{sec:cg1}. If more accuracy is desired, these initial candidates can be refined by matching other contour fragments near the candidates to other parts of the PCA model,  as described in Section \ref{sec:cg2}. 
\vskip -3mm
\begin{figure}[ht]
\centering
\includegraphics[width=4.1cm]{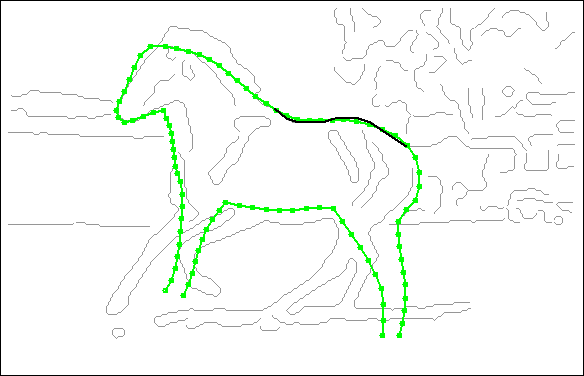}
\includegraphics[width=4.1cm]{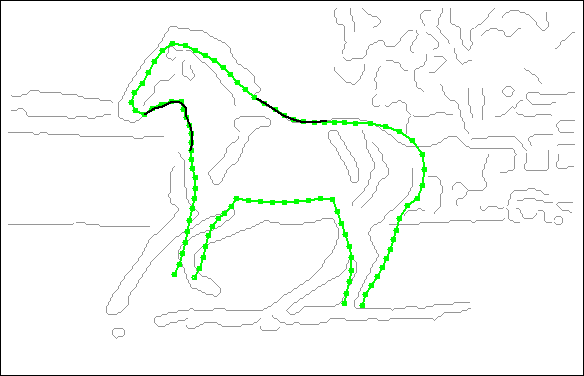}
\vskip -0mm
\caption{The best candidate obtained from one (left) and two (right) contour fragments. The fragments that generated each candidate are shown in black.}
\label{fig:cand_cg}
\vspace{-5mm}
\end{figure}

\subsubsection{Candidate Generation from One Contour Fragment}\label{sec:cg1}

These PCA candidates are obtained by matching a contour fragment to different parts of the PCA model. A non-maximal suppression step is performed to obtain candidates that are different from each other, since too similar candidates are likely to end up in the same local energy optimum.

To speed-up computation, an interval $[L(l),U(l)]$, representing the number of PCA points that match contour fragments of length $l$ (rounded to the nearest integer), is obtained from the training set and the ground truth annotations. 

The whole procedure is described in Algorithm \ref{alg:cg1}.
\vskip -3mm
\begin{algorithm}[htb]
   \caption{{\bf CG1}($N^{cand}$)}
   \label{alg:cg1}
\begin{algorithmic}
   \STATE {\bfseries Input:} Contour fragments $c$ of length $len(c)\in [l_{min},l_{max}]$
   \STATE {\bfseries Output:} At most $N^{cand}_1$ different PCA shape candidates $(A_i,\beta_i)$ with matches $(c_i,b_i,k_i)$.
 	 \FOR { any contour fragment $c$} 
 	 \FOR { any $k$ with $L(l)\leq k\leq U(l)$ where $l=[len(c)]$}
    \STATE Subsample $c$ evenly to have $k$ points $p_1,...,p_k$.
	 \FOR { $1\leq b\leq N$}
   \STATE Fit points $b,...,b+k-1$ of PCA shape $(A,\beta)$ to $p_1,...,p_k$ in a least square sense.
   \STATE Discard $(A,\beta)$ if the matching error is above a threshold.
   \ENDFOR
   \ENDFOR
   \ENDFOR
   \STATE Perform Non-Max Suppression to keep at most $N^{cand}_1$ candidates.
\end{algorithmic}
\end{algorithm}
\vskip -3mm

The Weighted PCA \cite{rogers2006robust}, described in the Appendix, is used to fit in a least square sense a given subset of a PCA shape to a number of points $p_1,...,p_k$.

The non-maximal suppression step finds the candidate of smallest fitting error and removes all candidates at average point-to-point distance at most $D^{nms}_1$ from it, then adds the remaining candidate of smallest error, and so on.

For each obtained PCA shape candidate $(A_i,\beta_i), i=1,...,N^{cand}_1$ we also remember the contour fragment $c_i$ and match location $(b_i,k_i)$ that were used to generate it.

In Figure \ref{fig:cand_cg}, left is shown the closest candidate to the ground truth among $N^{cand}_1=200$ candidates obtained by Algorithm \ref{alg:cg1}.

\subsubsection{Candidate Generation from Two Contour Fragments}\label{sec:cg2}

Usually images contain more than one contour fragment of  the object to be segmented. We can refine a candidate obtained by {\bf CG1} by fitting it simultaneously to the contour fragment it was obtained from and to another fragment close to the shape. The details of this strategy are given in Algorithm \ref{alg:cg2}. 
In Figure \ref{fig:cand_cg}, right is shown the closest candidate to the ground truth among $N^{cand}_2=400$ candidates obtained by Algorithm 
\ref{alg:cg2}.
Experiments in Section \ref{sec:results} show that {\bf CG2} can improve the quality of the candidates and of the final result. 
\begin{algorithm}[htb]
   \caption{{\bf CG2}($N^{cand}$)}
   \label{alg:cg2}
\begin{algorithmic}
   \STATE {\bfseries Input:} PCA shape candidates $(A_i,\beta_i)$ with matches $(c_i,b_i,k_i)$ from {\bf CG1} and contour fragments $c$.
   \STATE {\bfseries Output:} At most $N^{cand}$ different PCA shape candidates $(A_i,\beta_i)$.
 	 \FOR { $i = 1$ {\bf to} $N^{cand}_1$ }
 	 \STATE Set $(P_1,...,P_N)'=S(A_i,\beta_i)$ from Eq. \eqref{eq:pca}.
 	 \FOR { any contour fragment $c$}
 	 \STATE Find $P_j,P_k, 1\leq j,k\leq N$ closest to the beginning and end of $c$
 	 \IF{ $d(c,P_j)+d(c,P_k)<2d^{max}$ and $[j,k]$ does not overlap with $[b_i,b_i+k_i-1]$}
   \STATE Subsample $c$ to have $m=k-j+1$ points $p_1,...,p_m$.
   \STATE Find PCA shape $(A,\beta)$ that fits points $b_i,...,b_i+k_i-1$ through $c_i$ and $j,...,k$ through $p_1,...,p_m$ in a least square sense.
   \STATE Discard $(A,\beta)$ if the matching error is above a threshold.
   \ENDIF
   \ENDFOR
   \ENDFOR
   \STATE Perform Non-Max Suppression to keep at most $N^{cand}$ candidates.
\end{algorithmic}
\end{algorithm}


\subsection{Preprocessing} \label{sec:preprocessing}

\begin{figure}[ht]
\centering
\includegraphics[width=4.1cm]{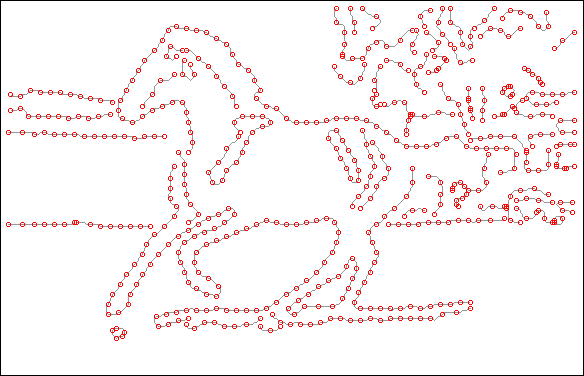}
\includegraphics[width=4.1cm]{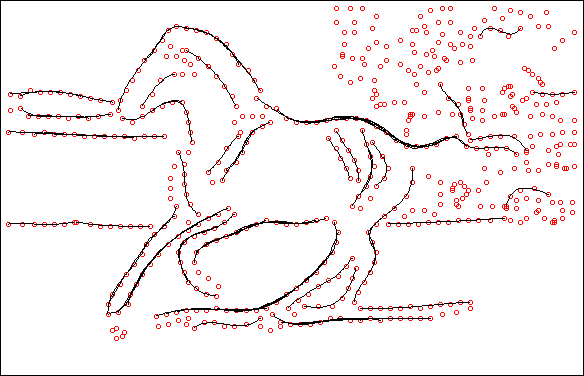}
\vskip -0mm
\caption{Left: the input points are traced into point chains and subsampled every 5-6 pixels. Right: smooth contour fragments (black) are fitted through the point chains starting and ending in the subsampled pixels.}
\label{fig:curvepts}
\vspace{-3mm}
\end{figure}

Preprocessing begins with tracing the input points into point chains based on the 8-neighborhood. The point chains are then subsampled every 5-6 pixels to reduce the number of contour fragments obtained, as illustrated in Figure \ref{fig:curvepts}, left.
\vskip -2mm

\begin{figure}[ht]
\centering
\includegraphics[width=5.cm]{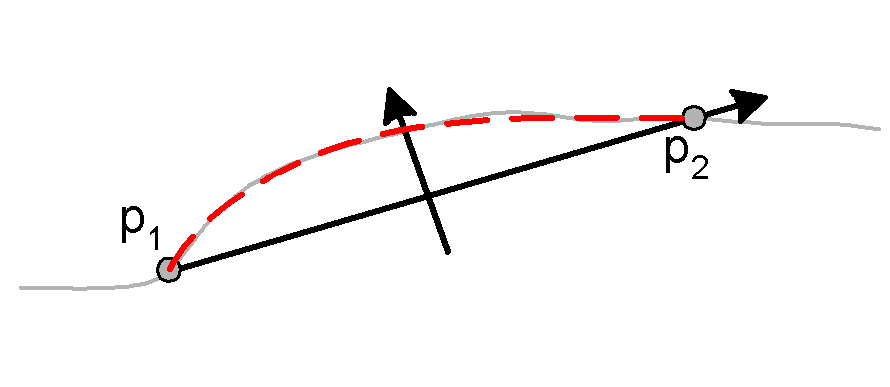}
\vskip -1mm
\caption{A contour fragment (red dashed) is a polynomial fit of a subset of a chain of points (shown in gray).}
\label{fig:curve}
\end{figure}

The contour fragments used by the candidate generators are represented as a polynomials of degree three relative to a system of coordinates aligned with the contour's endpoints, as illustrated in Figure \ref{fig:curve}.  
\vskip -2mm
\vskip -3mm

The contour fragment endpoints are two of the subsampled points of the same traced point chain and the polynomial is fitted in a least square sense through all the chain points in between. The fragments are restricted in length to a range $[l_{min},l_{max}]$. Only the fragments with a maximum error at most $e_{max}=1.5$ pixels are kept. 

The contour fragments obtained this way have a partial order inherited from the partial order between the sets of chain points they were constructed from.
\vskip -4mm
\begin{figure}[ht]
\centering
\includegraphics[width=5.cm]{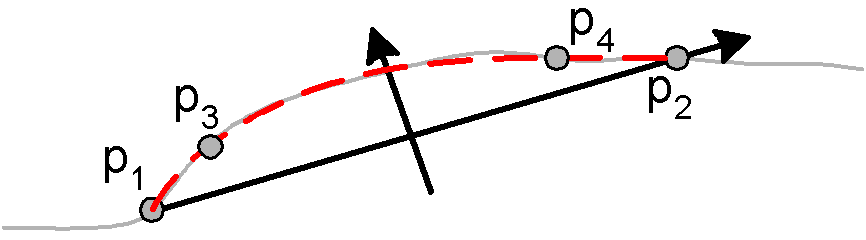}
\vskip -0mm
\caption{The contour fragment between $p_3$ and $p_4$ is a subset of the contour fragment between $p_1$ and $p_2$, i.e.  $\overline{p_3p_4}\subset \overline{p_1p_2}$.}
\label{fig:subcurve}
\vspace{-4mm}
\end{figure}

For example $\overline{p_3p_4}\subset \overline{p_1p_2}$ in Figure \ref{fig:subcurve}.
Based on this partial order, non-maximal contour fragments (e.g. $\overline{p_3p_4}$ from Figure \ref{fig:subcurve}) are removed.

An example of obtained contour fragments is shown in Figure \ref{fig:curvepts}, right.

\subsection{Learning the Model and Algorithm Parameters}\label{sec:learn}

The proposed model is very simple. It consists of a PCA model that in practice has at most 10 principal directions plus a small number ($<$20) of parameters.
Because the model is small, it should be expected that it generalize well to unseen data if the training data is representative.

The PCA model is learned in the standard way using Procrustes analysis to align the training shapes.

To learn the rest of the parameters, we adopt a supervised approach previously used successfully in other MRF-based methods \cite{barbu2009training,li_learning_2008,scharstein_learning_2007}, namely learning the parameters by optimizing a loss function on the training set. To speed-up the parameter learning, for candidate generators we employ loss functions that directly evaluate the generated candidates instead of the final result. 

The parameters of the candidate generators are learned first, in the order {\bf CG1} and {\bf CG2}, using the minimum of the average point-to-point distances from the candidates to the ground truth annotation (described in Section \ref{sec:results}) as loss function. This speeds-up the learning process since the {\bf CG} parameters are this way decoupled from the later modules. Other measures, such as detection rate/false positive rate for the contour fragments, could be used instead and are subject to further investigation. The number of PCA components was fixed to $p=4,8$ for {\bf CG1} respectively {\bf CG2} except for the faces where we used $p=2,4$ for {\bf CG1} respectively {\bf CG2}.

Similar to \cite{barbu2009training}, we adopted a coordinate descent optimization of the loss function, where at each step one parameter is perturbed by an increment and the change is kept if the loss function decreases. For {\bf CG1} and {\bf CG2} we restricted the number of candidates to balance speed and accuracy. The obtained parameters for {\bf CG1} are $l_{min}=20,l_{max}=60,N^{cand}_1=200,D^{nms}_1=5$,  and $N^{cand}_2=400,D^{nms}_2=8, d^{max}=20$ for {\bf CG2}. 

\begin{figure}[ht]
\centering
\includegraphics[width=4.1cm]{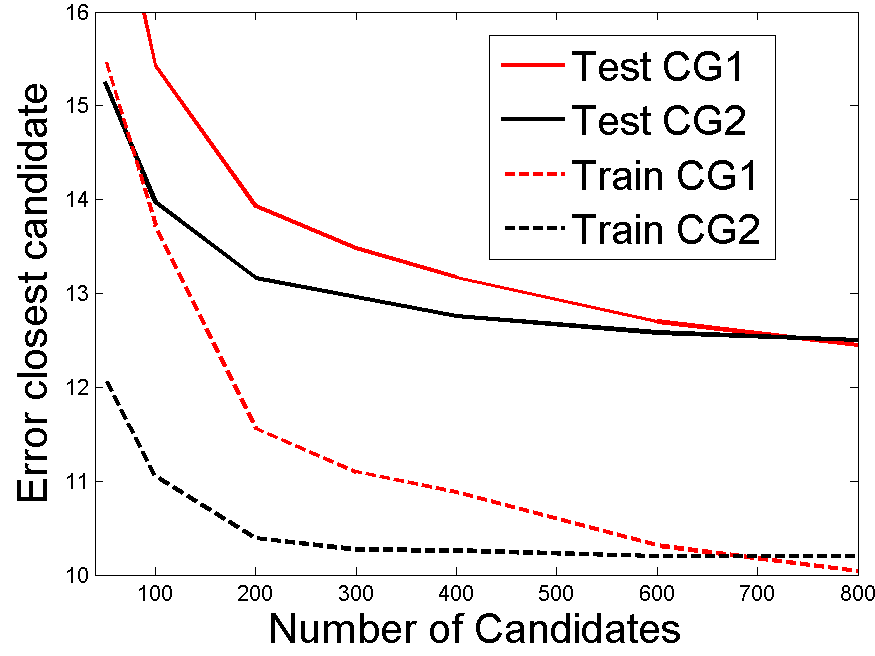}
\includegraphics[width=4.1cm]{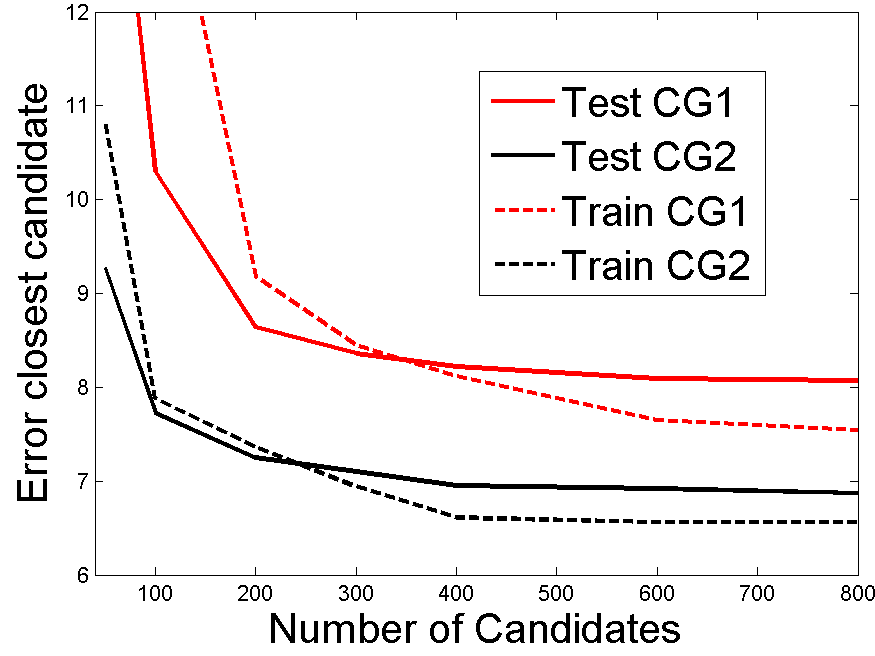}
\vskip -1mm
\caption{Candidate generator error vs. number of candidates for the horse (left) and cow (right) datasets.}
\label{fig:err_ncand}
\vspace{-3mm}
\end{figure}

In Figure \ref{fig:err_ncand} are shown the average errors of the closest candidate obtained by {\bf CG1-2} on the training and test sets vs the number of candidates $N^{cand}$. 
\vskip -2mm

Figure \ref{fig:err_ncand} shows that the test error decreases as the the training error decreases, which means that there is minimal overfitting for the candidate generators. Similar behaviors were observed for the $D^{nms}_1, D^{nms}_2$ parameters and for the $d^{max}$ parameter of {\bf CG2}. 
It is worth noting that the candidates from {\bf CG2} are usually better  than those of {\bf CG1} (closer to the ground truth), and the difference is larger for the cow images.

The model parameters $\Theta=(\alpha,\gamma,\delta,\rho,r)$ are learned based on the average point-to-point distance between the obtained parsing results and the ground truth annotation. 
\vspace{-2mm}
\begin{equation}
	Err(\Theta)=\frac{1}{n}\sum_{i=1}^n err_i(\Theta) \label{eq:err}
\vspace{-1mm}
\end{equation}
where $err_i(\Theta)$ is the average point-to-point error of the parsing result obtained with parameters $\Theta$ on example $i$ using $\bf CG1$. 
\vskip -3mm
 \begin{table}[htb]
\small 
\begin{center}
\caption{Learned parameters for Algorithm \ref{alg:inference}.}\label{tab:params}
\begin{tabular}{|l|c|c|c|c|c|}
\hline
Dataset \phantom{$^I$}& $\alpha_i$ &$\delta$ &$\rho$ &$r$ &$p$\\
\hline
\hspace{-1mm}Weizmann horses  \cite{borenstein2002class}\hspace{-1mm}\phantom{$^I$} &0.04 &2 &2 &1 &10\\
\hspace{-1mm}Cows  \cite{leibe2004combined}\hspace{-1mm} &0.04 &2 &2 &0.5 &10\\
\hspace{-1mm}IMM Faces  \cite{IMM20020922,Stegmann2003tmi}\hspace{-1mm} &0.04 &6 &2 &0.5 &5\\
\hline
\end{tabular}
\end{center}
\vspace{-4mm}
\end{table}
 \vskip -5mm
\begin{figure}[ht]
\centering
\includegraphics[width=4.1cm]{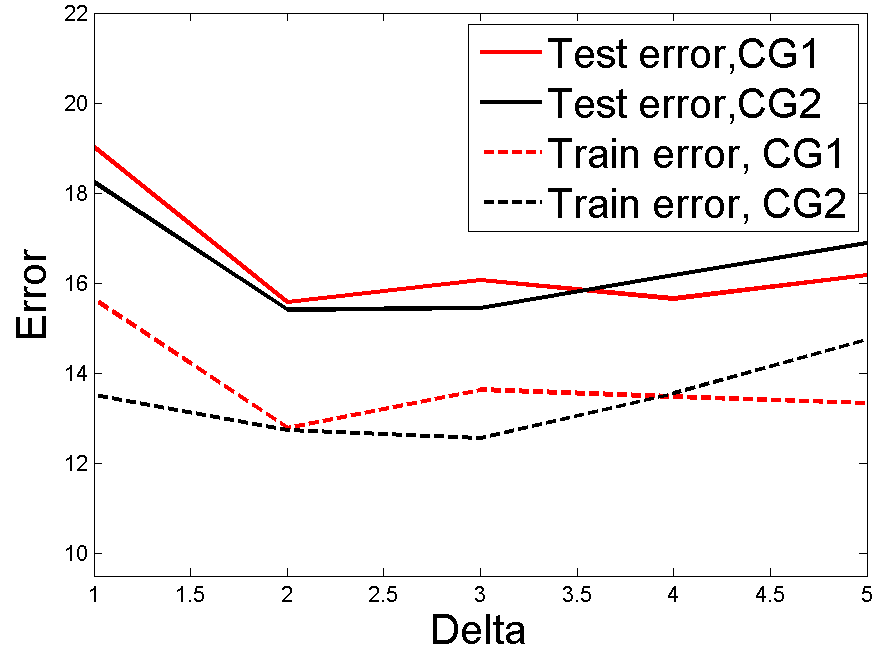}
\includegraphics[width=4.1cm]{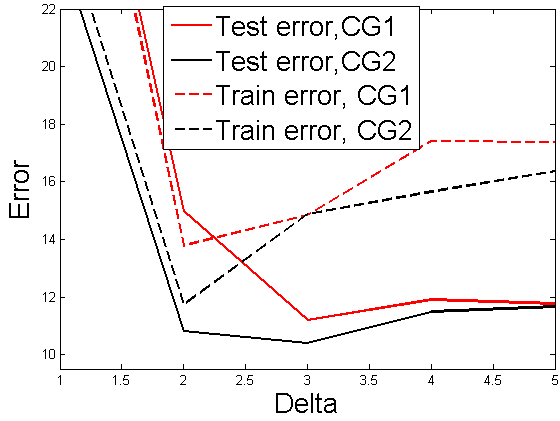}
\vskip -1mm
\caption{The parsing error measure \eqref{eq:err} vs $\delta$ for the horses (left) and cows (right).}
\label{fig:err_rho}
\vspace{-3mm}
\end{figure}
 \begin{table*}[htb]
\small 
\begin{center}
\caption{Performance of different methods on the Weizmann Horse dataset.}\label{tab:results}
\vskip 0.5mm
\begin{tabular}{|l|c|c|c|c|c|c|}
\hline
Method & Train &Test &Contour &Train&Test&Time/img\\
			&images&images&points &error &error &(sec)\\
\hline
\hspace{-1mm}Active Shape Model \cite{cootes1995active}\phantom{$^I$} \hspace{-3mm} &50  &227 &96 &25.35 &29.05 &$<$1\\
\hspace{-1mm}Recursive Compositional Models \cite{zhu2009learning} \hspace{-1mm} &1 &227 &27 &- &18.7 &3\\
\hspace{-1mm}Recursive Compositional Models \cite{zhu2009learning} \hspace{-1mm} &50 &227 &27 &- &16.04 &23\\
\hspace{-1mm}Ours, with CG1 \hspace{-1mm} &50 &227 &96 &12.79 &15.58 &44\\
\hspace{-1mm}Ours, with CG2 \hspace{-1mm} &50 &227 &96 &{\bf 12.74} &{\bf 15.36} &69\\
\hspace{-1mm}Ours, with CG2, no head or legs \hspace{-1mm} &50 &227 &60 &8.21 &11.42 &20\\
\hline
\end{tabular}
\end{center}
\vspace{-5mm}
\end{table*}
 \begin{table*}[htb]
\small 
\begin{center}
\caption{Performance of different methods on the Cows dataset.}\label{tab:resultscow}
\vskip 0.5mm
\begin{tabular}{|l|c|c|c|c|c|c|}
\hline
Method & Train &Test &Contour &Train&Test&Time/img\\
			&images&images&points &error &error &(sec)\\
\hline
\hspace{-1mm}Active Shape Model \cite{cootes1995active} \phantom{$^I$} \hspace{-3mm} &25  &111 &87 &48.81 &49.23 &$<$1\\
\hspace{-1mm}Recursive Compositional Models \cite{zhu2009learning} \hspace{-1mm} &1  &111 &27 &- &15.8 &3.5\\
\hspace{-1mm}Ours, with CG1\hspace{-1mm} &25  &111 &87 &13.78 &14.98 &14\\
\hspace{-1mm}Ours, with CG2\hspace{-1mm} &25  &111 &87 &{\bf 11.73} &{\bf 10.81} &28\\
\hline
\end{tabular}
\end{center}
\vspace{-5mm}
\end{table*}
\begin{table*}[htb]
\small 
\begin{center}
\caption{Performance of different methods on the IMM Face dataset at $320\times 240$ resolution.}\label{tab:resultsimmface}
\vskip 0.5mm
\begin{tabular}{|l|c|c|c|c|c|c|c|c|}
\hline
Method & Number of &Uses &Automatic&Crossval.  &Contour &Train &Test&Time/img\\
			&images &intensity &initialization &folds &points &error &error &(sec)\\
\hline
\hspace{-1mm}Active Shape Model \cite{cootes1995active}\phantom{$^I$} \hspace{-3mm} &40 &No &Yes &4 &58 &21.47 &21.56 &0.08\\
\hspace{-1mm}Stegman \cite{Stegmann2003tmi} \hspace{-1mm} &37 &B/W &No &37 &58 &- &3.14 &0.13\\
\hspace{-1mm}Stegman \cite{Stegmann2003tmi} \hspace{-1mm} &37 &Color &No &37 &58 &- &{\bf 3.08} &0.28\\
\hspace{-1mm}Ours, with CG1\hspace{-1mm} &40 &No &Yes &4 &58 &6.54 &6.64 &0.33\\
\hspace{-1mm}Ours, with CG2\hspace{-1mm} &40 &No &Yes &4 &58 &5.30 &5.57 &0.43\\
\hline
\end{tabular}
\end{center}
\vspace{-8mm}
\end{table*}

The model parameters $\delta,\rho,\alpha_i=\alpha,r,p$ were obtained by optimizing the error measure \eqref{eq:err} on the training set by coordinate descent, with fixed parameters $\gamma_i=0.1$ (to fix the scale of the energy function \eqref{eq:model}) and $N_{iter}=10$. 
The obtained values are given in Table \ref{tab:params}. 

The dependence of the error on the values of $\delta$ is shown in Figure \ref{fig:err_rho} for the horses and cows. Again, the test errors follow the training error. 

\section{Experimental Results}\label{sec:results}

We evaluated this model and algorithm combination on three datasets: the Weizmann dataset \cite{borenstein2002class} containing 328 horse images with object segmentations as  binary masks, the Cows dataset \cite{leibe2004combined} with 111 cow images and the IMM face dataset  \cite{IMM20020922,Stegmann2003tmi}. We used the same subsets of images as \cite{zhu2009learning} for training and testing the Weizmann dataset and the first 25 images for training the Cows dataset and tested on all 111 images.

Many works \cite{borenstein2008combined,cour2007recognizing,kumar2005obj,levin2009learning,ren2006cue,winn2005locus,zhu2009learning} report segmentation results on the Weizmann horses in terms of percentage of correctly classified pixels. However, object segmentation is a different problem than object parsing since a pixel segmentation has no information on the position of the object parts. Moreover, these works make use of the intensity information in different ways in obtaining the segmentation. We cannot do the same in our problem as we have no intensity information available from the edge detection images..
\begin{figure}[ht]
\centering
\includegraphics[width=4.1cm]{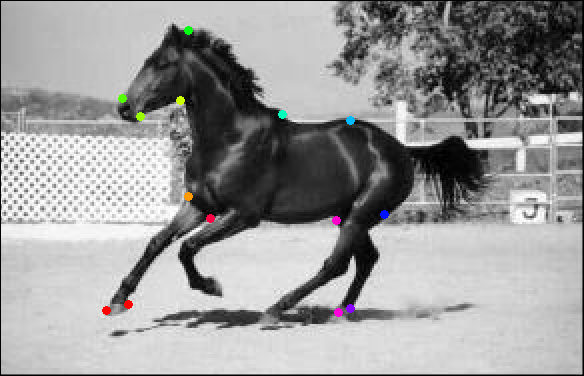}
\includegraphics[width=4.1cm]{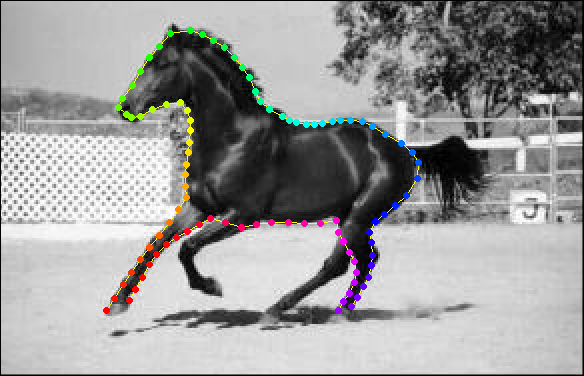}
\includegraphics[width=4.1cm]{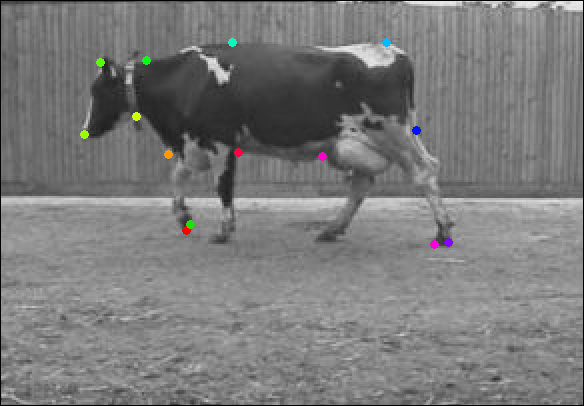}
\includegraphics[width=4.1cm]{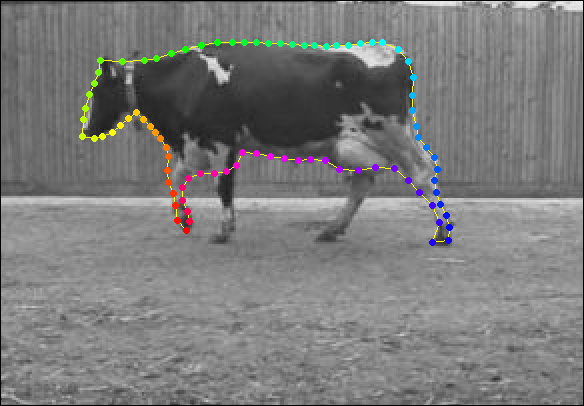}
\vskip -0.mm
\caption{Left: The horses and cows are annotated with 14 control points on the contour and the two outer legs. Smooth curves (yellow) are obtained between the control points using the ground truth segmentation. Right: The obtained boundary annotation with 96 points (horses) and 87 points (cows).}
\label{fig:ano}
\vspace{-8mm}
\end{figure}

Each horse and cow were manually annotated with 14 control points on the boundary, as illustrated in Figure \ref{fig:ano}, left. For fairness of comparison, the same horse and cow legs (usually the outer legs) were annotated as in \cite{zhu2009learning}. This is not an extreme case. If the legs for annotation were chosen at random, there would be larger shape variability and a decreased performance. If the legs were chosen to minimize the shape variability, a better performance could be obtained.

\begin{figure*}[ht]
\centering
\hspace{-1mm}\includegraphics[width=8.5cm]{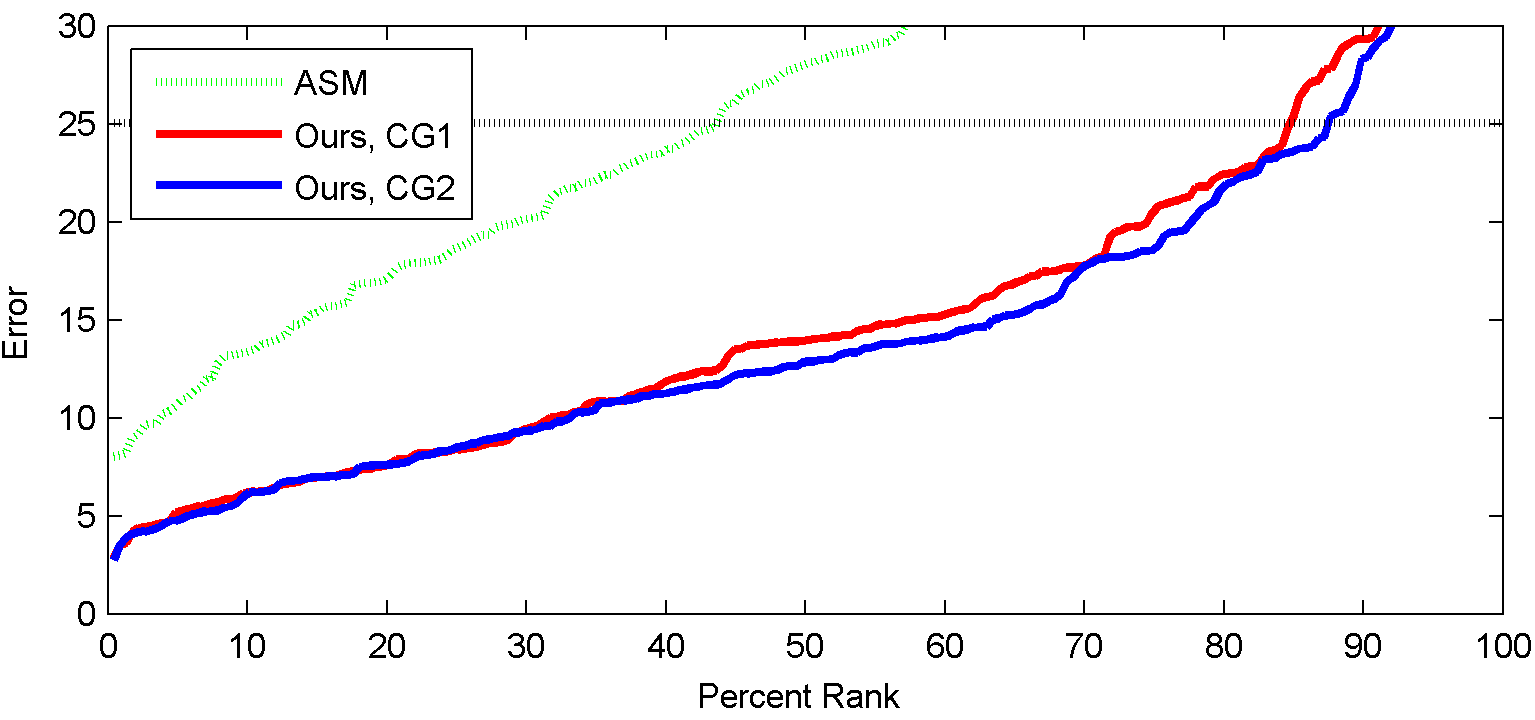}
\hspace{-1mm}\includegraphics[width=8.5cm]{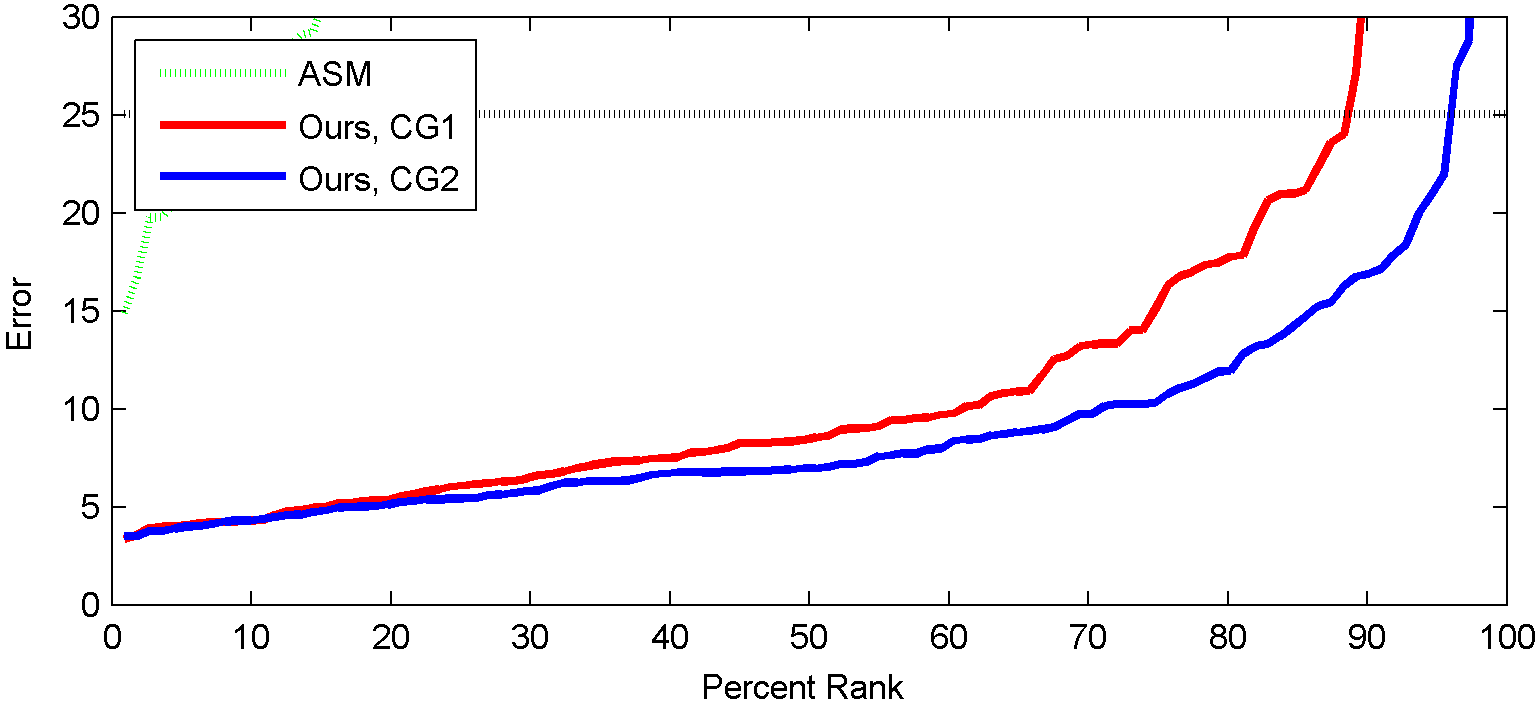}
\hspace{-1mm}\includegraphics[width=8.5cm]{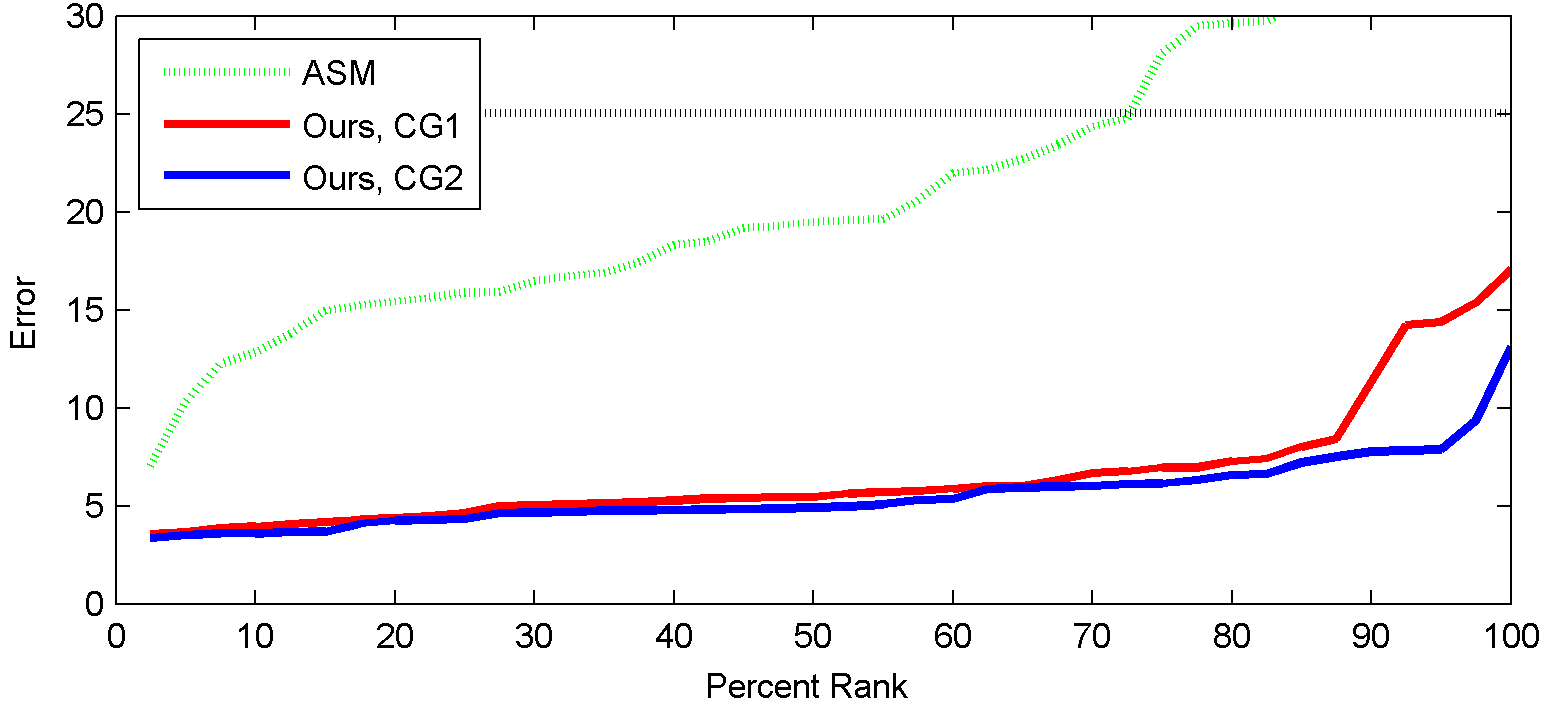}
\vskip -1mm
\caption{The sorted errors \eqref{eq:err} of our algorithm on the 227 test images from the Weizmann dataset (top left), on the 111 images of the cow dataset (top right) and on the 40 frontal images of the IMM face dataset (bottom).}
\label{fig:ploterr}
\vspace{-6mm}
\end{figure*}

Smooth curves were obtained between the control points by dynamic programming to minimize the average distance to the object boundary from the binary mask. Intermediate points were obtained by dividing the smooth curves into equal parts. Examples of obtained annotations are shown in Figure \ref{fig:ano}, right, with 96-points for the horses and 87 points for the cows.

We also evaluated a standard Active Shape Model \cite{cootes1995active} initialized in the center of the image with average scale, no rotation $\theta=0$ and 20 update iterations.

 The results are summarized in Tables \ref{tab:results}, \ref{tab:resultscow} and  \ref{tab:resultsimmface}. In   Fig. \ref{fig:ploterr} are plotted the sorted errors on the datasets, from which different error percentiles can be obtained. The Recursive Compositional Model  \cite{zhu2009learning} also reports average point-to-point errors on the Weizmann and cow datasets but uses both edge and intensity information, unlike our approach, which only uses edge information. 

The second horse result from \cite{zhu2009learning} uses a model trained on 50 images manually annotated with  27-point contours similar to ours. The first horse result from \cite{zhu2009learning} and the cow result are based on models trained on a single image without annotation, which are not as good as models trained with manual annotation. 

The face model is evaluated on the IMM face dataset \cite{IMM20020922,Stegmann2003tmi}  containing 240 face images of 40 people, each with six different poses, varying illumination and expressions. The AAM model from \cite{Stegmann2003tmi} is evaluated on 37 out of the 40 frontal faces (probably excluding the three grayscale images). We evaluate our model on all 40 frontal faces of the dataset with four fold cross-validation. The results are summarized in  \ref{tab:resultsimmface}. The best AAM errors from  \cite{Stegmann2003tmi} are 3.08 using color information while our algorithm with {\bf CG2} obtains 5.57. However, our algorithm does not use any intensity information and is fully automatic, whereas  \cite{Stegmann2003tmi} initializes their algorithms at locations close to the true location and report the error after convergence. The argument is that a face detector can be used to initialize the AAM. However, both the face detector and the AAM use intensity information, so they cannot work on the edge detection images.  We are not aware on any fully automatic face alignment results on this dataset. 

The Recursive Compositional Model \cite{zhu2009learning} reports 6 pixel average point-to-point errors on the face dataset  \cite{li2005robust}, so comparable results should be expected for the IMM face dataset. 

One factor that greatly influences performance is our data term that works best when the contour points are close to each other (about 5 pixels or so), because the further the contour points are from each other, the more likely it is that the edge chain is broken between consecutive points. This disadvantage could be alleviated by  increasing the density of the contour points, as we did for the horses and cows, or by using a more elaborate data term that also considers the gaps between contours. 

The PCA model has difficulties modeling the shape variability of the horse head and legs. If the head and leg points are removed both from the model and from the evaluation, the training and test errors decrease substantially, as it can be seen in the last row of Table \ref{tab:results}. This experiment suggests that using part-based shape models with free parameters for the head and leg positions might be more appropriate than the PCA for the higher level model. Such models are subject to further investigation. 

\begin{figure*}[htb]
\centering
\includegraphics[width=4.cm]{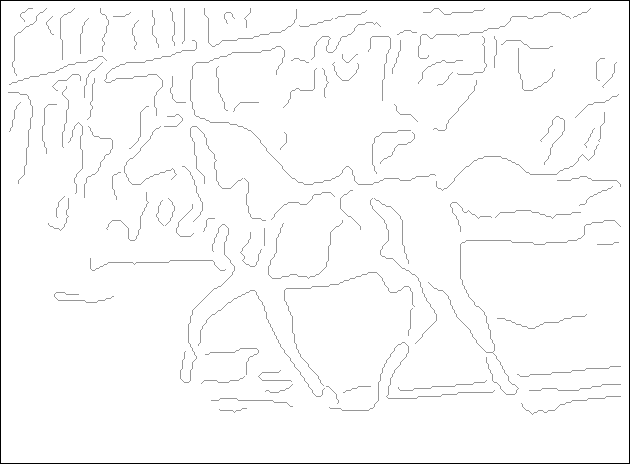}
\includegraphics[width=4.cm]{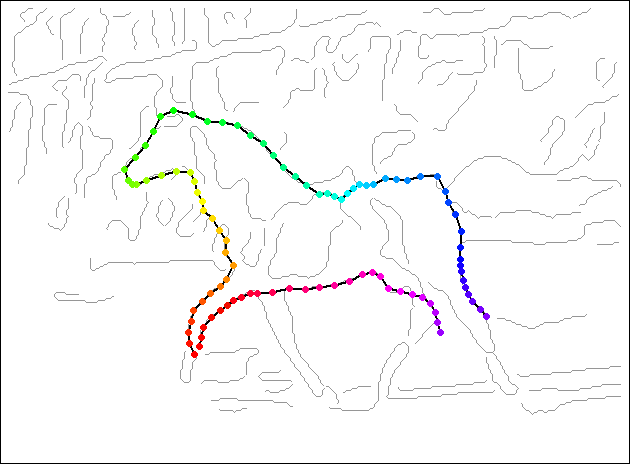}
\includegraphics[width=4.cm]{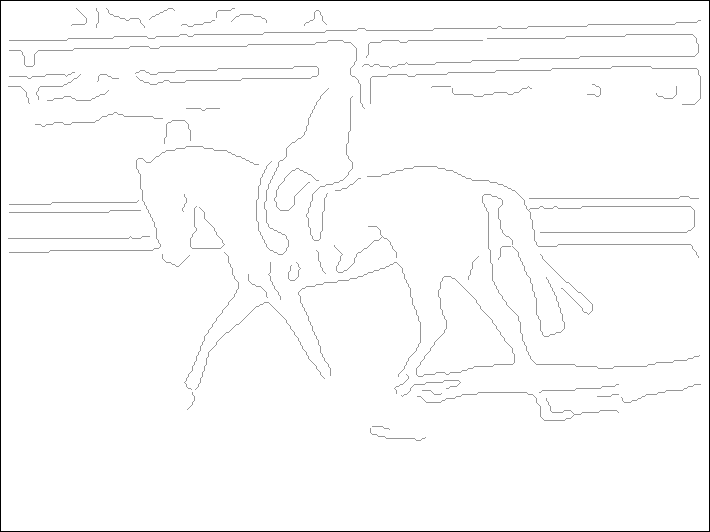}
\includegraphics[width=4.cm]{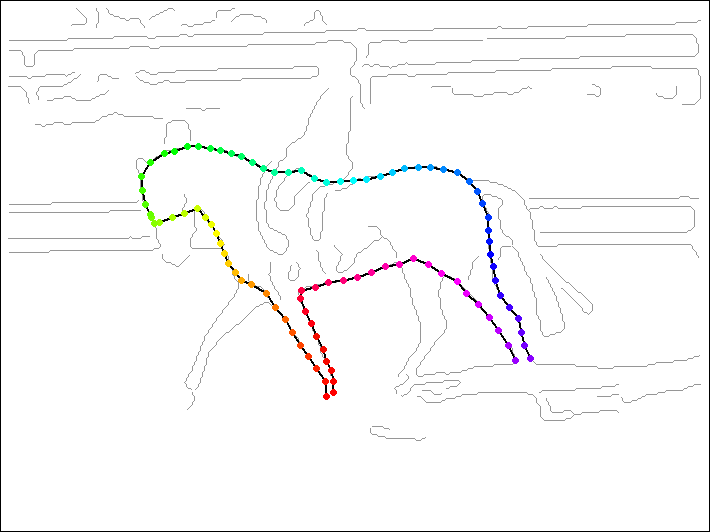}
\includegraphics[width=4.cm]{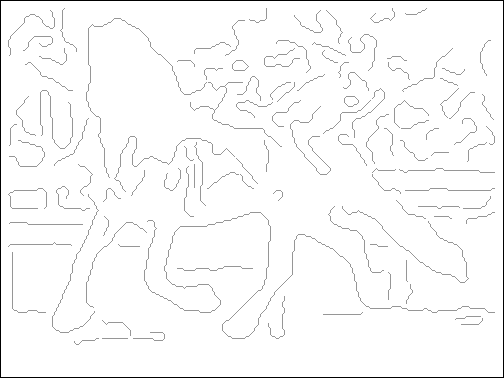}
\includegraphics[width=4.cm]{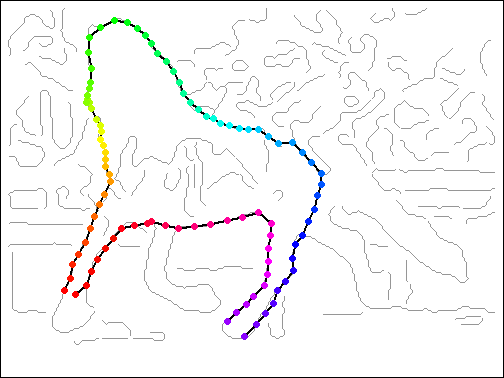}
\includegraphics[width=4.cm]{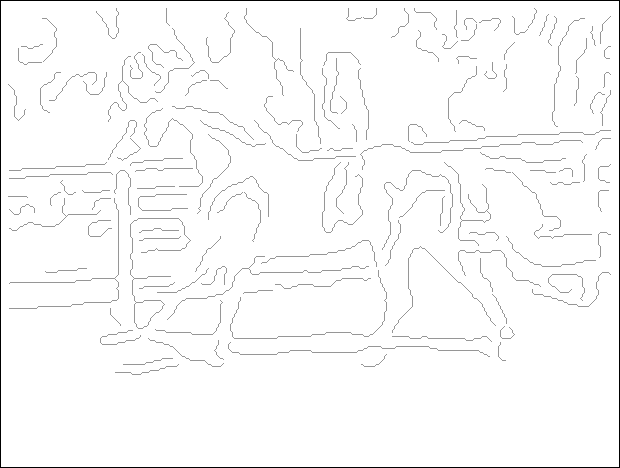}
\includegraphics[width=4.cm]{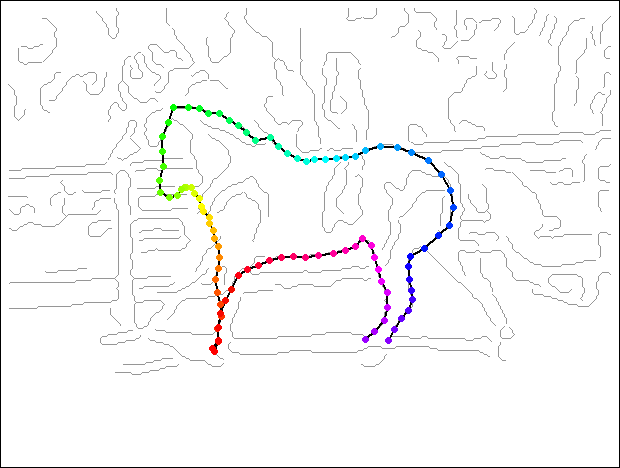}
\includegraphics[width=4.cm]{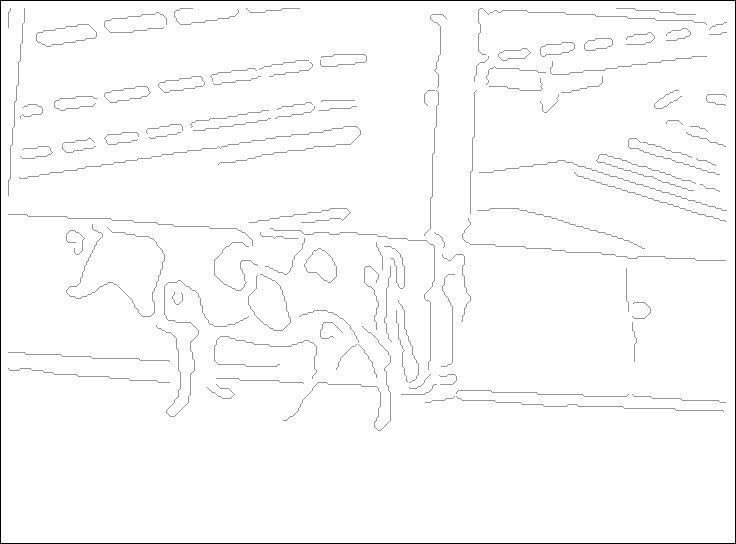}
\includegraphics[width=4.cm]{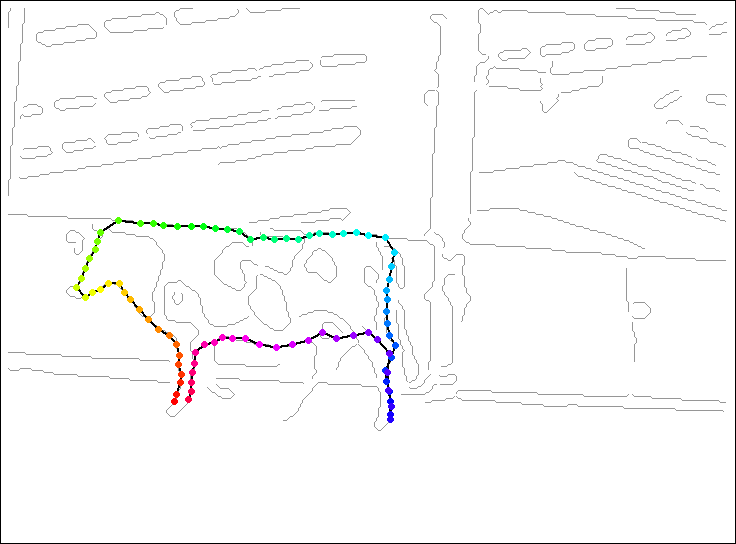}
\includegraphics[width=4.cm]{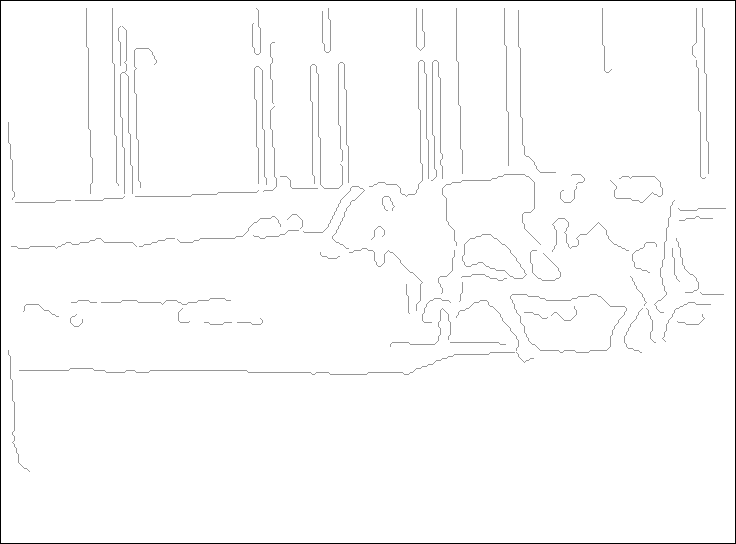}
\includegraphics[width=4.cm]{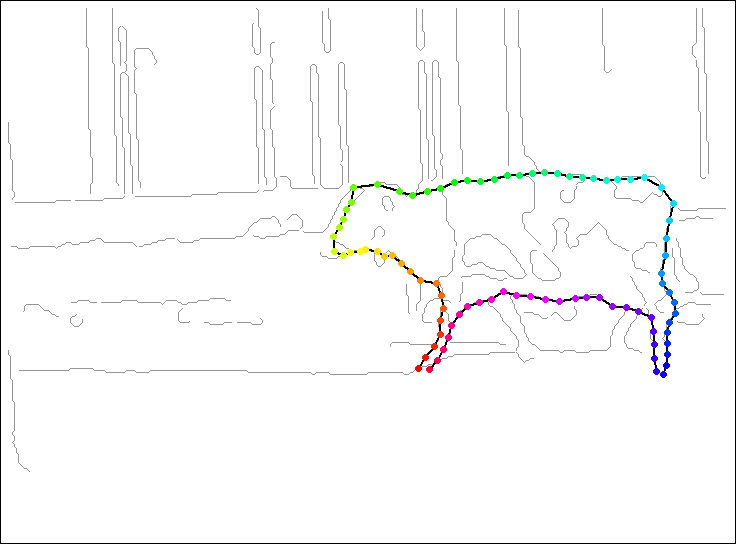}
\includegraphics[width=4.cm]{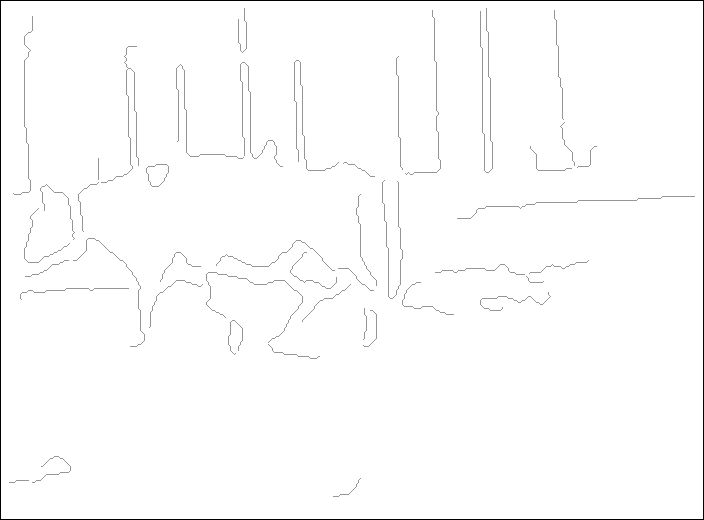}
\includegraphics[width=4.cm]{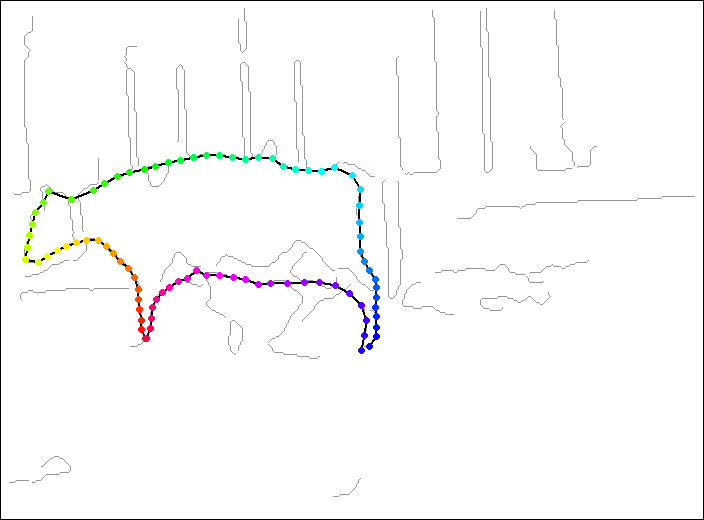}
\includegraphics[width=4.cm]{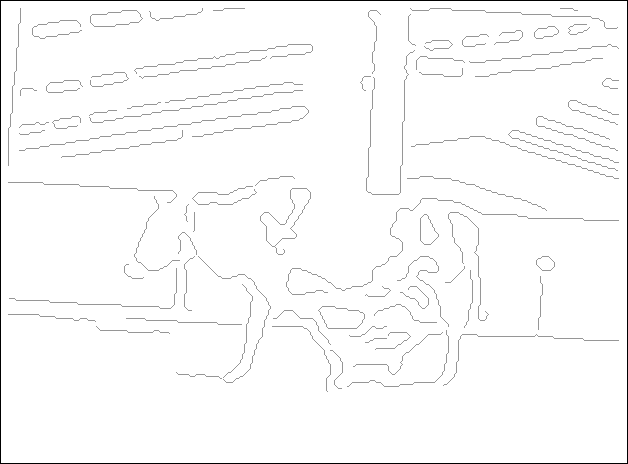}
\includegraphics[width=4.cm]{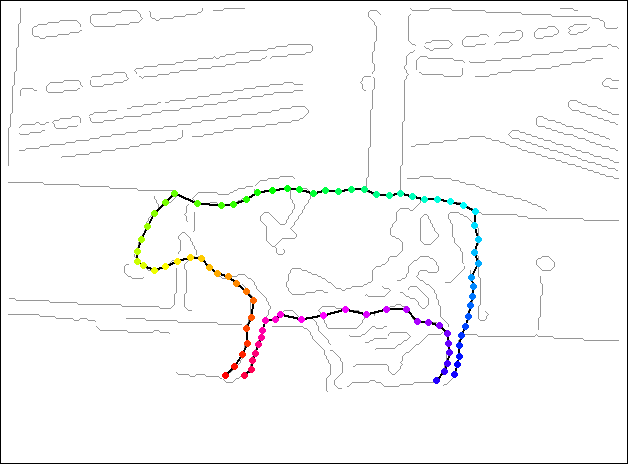}
\vskip -0mm
\caption{Results on the horse and cow datasets.}
\label{fig:resultshorsecow}
\vspace{-5mm}
\end{figure*}

\begin{figure*}[htb]
\centering
\includegraphics[width=4.0cm]{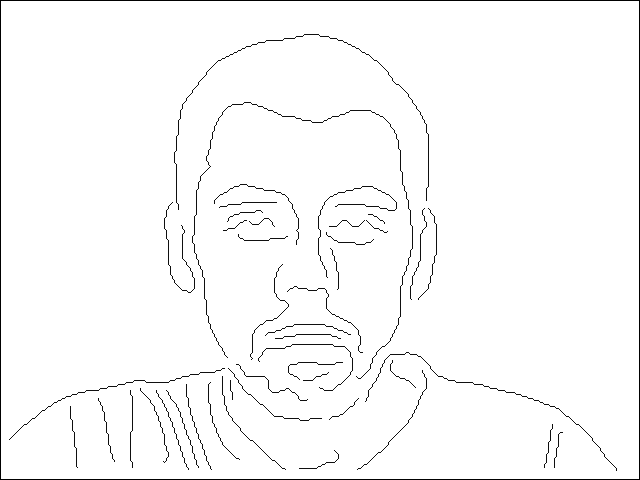}
\includegraphics[width=4.0cm]{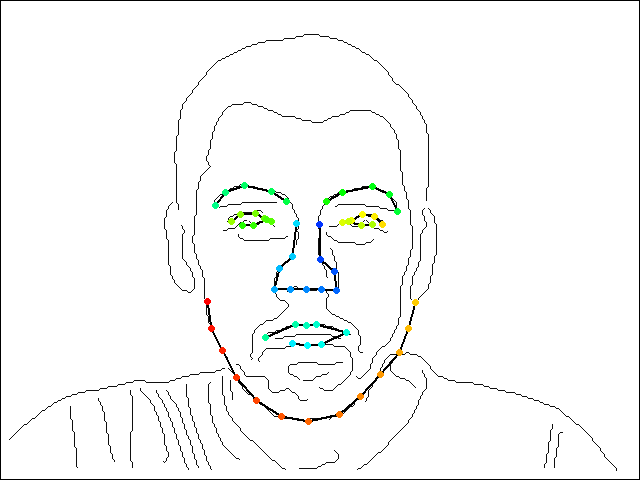}
\includegraphics[width=4.0cm]{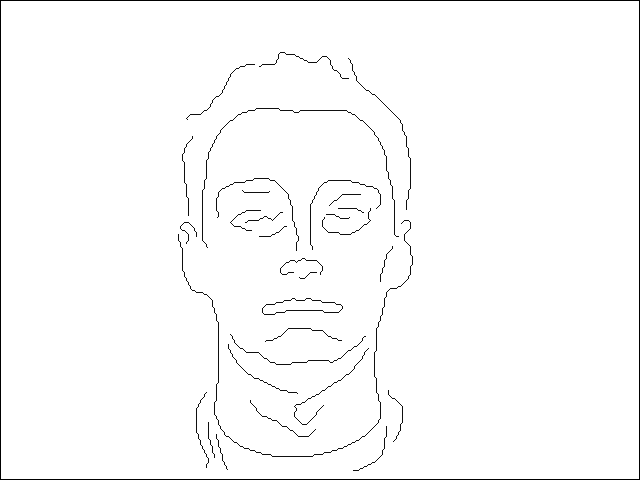}
\includegraphics[width=4.0cm]{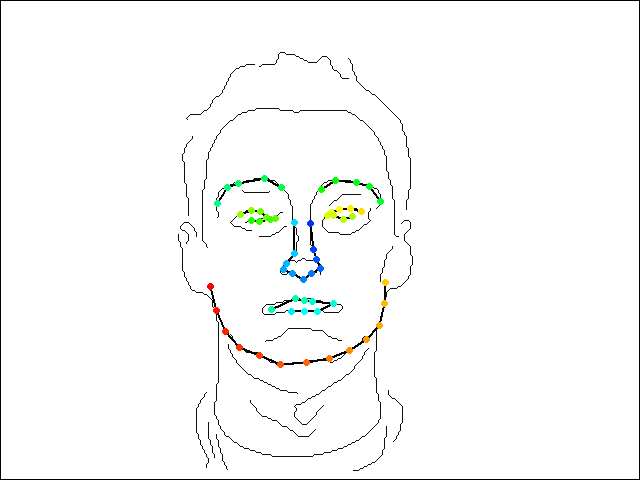}
\includegraphics[width=4.0cm]{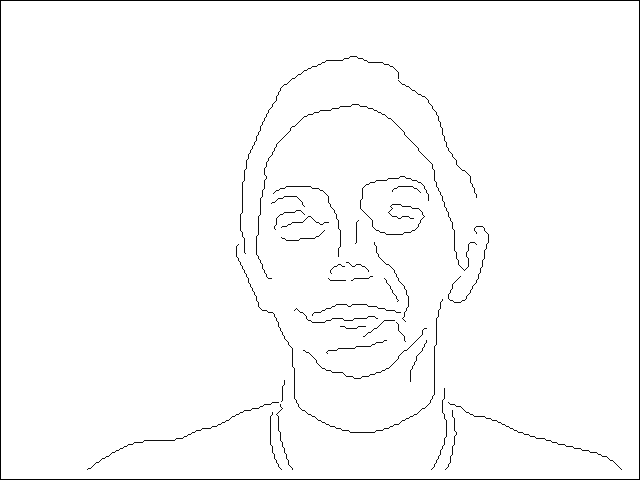}
\includegraphics[width=4.0cm]{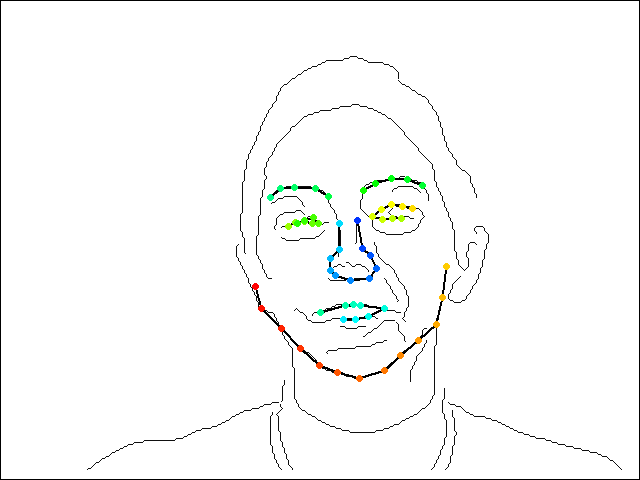}
\includegraphics[width=4.0cm]{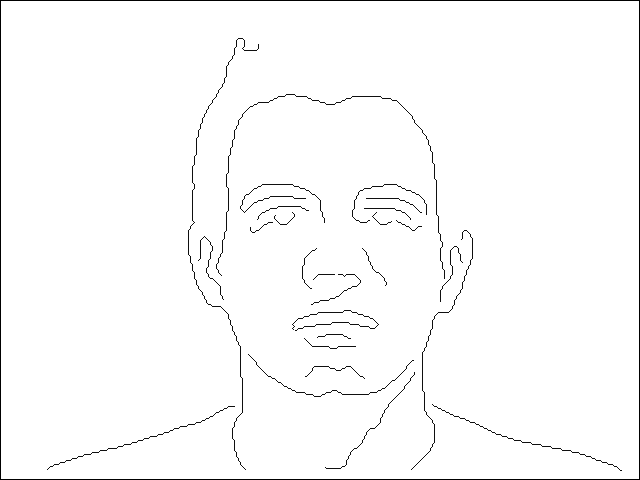}
\includegraphics[width=4.0cm]{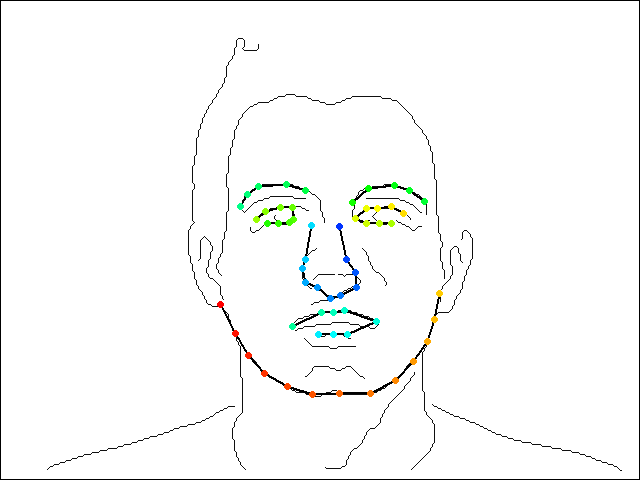}
\vskip -0mm
\caption{Results on the face dataset.}
\label{fig:resultsface}
\vspace{-5mm}
\end{figure*}

Parsing examples using {\bf CG2}, are shown in Figures \ref{fig:resultshorsecow} and \ref{fig:resultsface}. 

\section{Conclusion and Future Work}

This paper proposes a novel approach to object parsing and applies it to data coming from noisy point clouds such as edge detection images. The object shape is modeled as a MRF deformation of a hidden PCA model. The model energy is minimized through many local searches starting from a number of data-driven initializations. Based on the experimental evaluation we conclude that the proposed model is quite accurate, and even though the  inference algorithm is suboptimal, our method is competitive with modern approaches for object parsing from point clouds such as the Recursive Compositional Models \cite{zhu2009learning} and Active Shape Models \cite{cootes1995active}.

The candidate generators and the object parsing algorithm can be easily parallelized, expecting a 10-100 times speedup from a GPU implementation.

In the future, we plan to investigate more accurate three-level part-based models, with separate parameters for the head and leg shapes and positions. We also plan to extend the method to 3D object parsing using approximate inference methods such as Graph Cuts or Belief Propagation for the deformation inference. 

\vspace{-4mm}
{\small
\bibliographystyle{ieee}
\bibliography{SegmentPcabib}
}
\vspace{-3mm}

\section*{Appendix: Weighted PCA}
\vspace{-1mm}

A partial PCA model can be fit to a number of points using the weighted least squares method \cite{rogers2006robust}, summarized in Algorithm \ref{alg:wtasm}. The weights of missing PCA points are set to zero. The weighted alignment between the shapes $S_1,S_2$ has been described in 
Appendix A from \cite{cootes1995active}.
\vskip -3mm
\begin{algorithm}[htb]
   \caption{{\bf FitWeightedPCA}}
   \label{alg:wtasm}
   {\bfseries Input:} Shape $S_1=({\bf x}_1,{\bf y}_1)$, weight vector ${\bf w}=(w_1,...,w_N)'$, $\|{\bf w}\|_1=\sum_{i=1}^N w_i=1$.\\
   {\bfseries Output:} Weighted least-square fit parameters $(A,\beta)$\\
\vspace{-5mm}
\begin{algorithmic}
   \STATE Set $W=diag(w_1,...,w_N)$ 
   \STATE Set {\small $K_x=(P_x'W^2P_x)^{-1}P_x'W^2,K_y=(P_y'W^2P_y)^{-1}P_y'W^2$}
   \STATE Set $\beta=0$
 	 \FOR { $i=1$ {\bfseries to} $N_{it}$}
   \STATE Set {\small $S_2=({\bf x}_2, {\bf y}_2),\;{\bf x}_2=\mu_x+P_x\beta,\; {\bf y}_2=\mu_y+P_y\beta$}
   \STATE Solve
\vspace{-2mm}
{\small
\[
\begin{pmatrix}
S_{xx}+S_{yy} &0 &S_x &S_y\\
0 &S_{xx}+S_{yy} &-S_y &S_x\\
S_x &-S_y &S_w &0\\
S_y &S_x &0 &S_w
\end{pmatrix}\hspace{-2mm}
\begin{pmatrix}
a\\b\\dx\\dy
\end{pmatrix}\hspace{-1mm}=\hspace{-1mm}
\begin{pmatrix}
S_1\\
S_2\\
{\bf x}_2'{\bf w}\\
{\bf y}_2'{\bf w}
\end{pmatrix}
\vspace{-3mm} 
\]
with
\vspace{-5mm} 
\[
\begin{split}
&S_x={\bf x}_1'{\bf w}, S_y={\bf y}_1'{\bf w}, S_w=\|{\bf w}\|_1=1\\
&S_{xx}={\bf x}_1' W{\bf x}_1, S_{yy}={\bf y}_1' W{\bf y}_1\\
&S_1={\bf x}_1' W{\bf x}_2+{\bf y}_1' W{\bf y}_2\\
&S_2={\bf y}_1' W{\bf x}_2-{\bf x}_1' W{\bf y}_2
\end{split}
\vspace{-3mm}
\]
}
\vspace{-4mm} 
\STATE Obtain   $A({\bf x},{\bf y})=(a{\bf x}+b{\bf y}+dx, -b{\bf x}+a{\bf y}+dy)$
\STATE Find $({\bf x}_o,{\bf y}_o)=A^{-1}(S_1)$
\STATE Set $\beta=K_x({\bf x}_o-\mu_x)+K_y({\bf y}_o-\mu_y)$
   \ENDFOR
   \STATE Set $s=\sqrt{a^2+b^2}, \theta=\arctan (b/a)$.
   \STATE Obtain $A=(dx, dy, s,\theta)$
\end{algorithmic}
\end{algorithm}

\end{document}